\documentclass[aps,pre,superscriptaddress,amsmath,amssymb,nofootinbib]{revtex4-2}

\usepackage{amsthm}
\usepackage{graphicx}
\usepackage{booktabs}
\usepackage{multirow}
\usepackage{microtype}
\usepackage{enumitem}
\usepackage{caption}
\usepackage{bm}
\usepackage{subcaption}

\usepackage[colorlinks=false]{hyperref}
\usepackage{url}
\usepackage{xcolor}
\usepackage[normalem]{ulem}

\newif\ifmarked
\markedfalse
\ifmarked
  
  \newcommand{\del}[1]{\textcolor{red}{\sout{#1}}}
\else
  
  \newcommand{\del}[1]{}
\fi

\newcommand{\ket}[1]{|#1\rangle}
\newcommand{\bra}[1]{\langle #1|}
\newcommand{\braket}[2]{\langle #1|#2\rangle}
\newcommand{\Tr}{\mathrm{Tr}}

\newcommand{\GT}{\mathrm{GT}}
\newcommand{\lora}{\textsc{LoRA}}
\newcommand{\fullft}{\textsc{Full FT}}
\newcommand{\gmax}{G_{\max}}
\newcommand{\thetastar}{\theta^{*}}
\newcommand{\geff}{\mathcal{G}_{\rm eff}}
\newcommand{\lgap}{\zeta}

\begin{document}

\title{Phantom Transitions in Language Model Fine-Tuning: A Density-Matrix Analysis}

\author{Vaibhav Prakash}
\email{vaibhav.prakash@mahindrauniversity.edu.in}
\affiliation{Department of Physics, Mahindra University, Hyderabad, India}
\author{Jayasri Dontabhaktuni}
\email{jayasri.d@mahindrauniversity.edu.in}
\affiliation{Department of Physics, Mahindra University, Hyderabad, India}

\date{\today}

\begin{abstract}
Language models fine-tuned on contexts where the correct completion must outrank a near-synonym competitor often fail silently. The cross-entropy loss decreases monotonically. The correct token never overtakes the competitor in the model's ranking. We study this regime across five transformer architectures spanning two families and a sixfold parameter range. The test set is ten hand-selected contexts in which the correct completion and its nearest competitor share substantial embedding overlap. We construct an order parameter that combines the model's predicted distribution with the geometric overlap between token embeddings. A density-matrix construction is the natural framework for this order parameter, because the prediction distribution lives over a non-orthogonal basis of token embeddings and the geometry of that basis interferes with classical probability intuitions. The order parameter decomposes additively into two terms. The first is a signal that tracks the model's commitment to the correct token over its nearest competitor. The second is a background drag set by how the rest of the embedding bulk leaks probability into the score. This decomposition isolates two failure modes. In kinematic failure the signal stays too small and the model never commits. In structural failure the drag actively worsens as fine-tuning proceeds, so the model degrades geometrically even while its loss decreases. The order parameter also shows sharp jumps that resemble phase transitions. We test the natural spontaneous-symmetry-breaking interpretation of these jumps by tracking the order parameter after every gradient step, and we rule it out. The jumps persist on multiple sentences under LoRA fine-tuning even though the token embedding matrix stays unchanged across every training step. No geometric phase transition is possible when the embedding geometry cannot move, so the discontinuity lies entirely in the softmax readout. A small set of dimensionless quantities organizes the trajectory across architectures. One quantity is consistent across all five models under full fine-tuning. A second quantity sorts architectures into two classes by their bulk embedding distribution and predicts whether LoRA alone can drive a sentence to commit. As a blind test, the framework predicts the critical learning rate of a held-out architecture, not used to fit any parameter, to within 2.1\% of a subsequent learning-rate sweep. All quantitative findings are claims about the near-synonym geometric mechanism studied here, and they should not be extrapolated without re-calibration to general instruction-tuning datasets or broader task distributions.
\end{abstract}

\maketitle

\section{Introduction}
\label{sec:intro}

Fine-tuning a pre-trained transformer language model
\cite{vaswani2017attention,devlin2019bert,brown2020language,radford2019language,touvron2023llama,ouyang2022training}
minimises a cross-entropy objective against a gold-standard
next-token target. Its failure modes have been catalogued through
that same objective. These include seed
instability~\cite{mosbach2021stability}, small-data
variance~\cite{dodge2020finetuning}, feature
distortion~\cite{kumar2022fine},
and hallucination~\cite{ji2023survey,maynez2020faithfulness}. Parameter-efficient methods such as low-rank adaptation
\cite{hu2022lora,houlsby2019parameter,lester2021power,dettmers2023qlora,intrinsic2020}
are evaluated against the same yardstick.

A transformer language model predicts the next token through a fixed
pipeline, which we sketch here because it fixes the vocabulary this
paper builds on. Each input token is first mapped to a real vector,
its embedding, drawn from a shared embedding matrix. The collection of
these vectors, together with their relative directions, is what we
mean by the embedding geometry of a model. The sequence of embeddings
then passes through repeated layers combining a self-attention
mechanism~\cite{vaswani2017attention}, which lets the representation
at each position incorporate information from other positions in the
context, with local feed-forward transformations, producing a final
hidden state vector that summarises the model's prediction at the
current position. This hidden state is read out by a linear
projection, the LM head, onto one score per vocabulary token, called a
logit. The softmax function, the same normalised-exponential form as a
Boltzmann distribution, then converts these logits into a probability
distribution over the vocabulary.

Transformer embedding geometry has separately been shown to be
strongly anisotropic \cite{ethayarajh2019contextual,gao2019representation,cai2021isotropy},
with near-synonym tokens clustering well above the population-level
distribution of pairwise cosine similarities, which we call the bulk
and formalise in \S\ref{sec:universality}. The dynamical consequences
of this anisotropy for fine-tuning have received little direct
attention, particularly when the correct completion has a
geometrically close competitor.

We study one such regime. Training on contexts whose correct
completion has a near-synonym, for instance \emph{guilt} for \emph{shame},
\emph{rubble} for \emph{ashes}, or \emph{yearning} for \emph{longing},
often produces a monotonic loss decrease and a rising probability on the
correct token, yet the correct token never overtakes the competitor in
the model's ranking. Cross-entropy tracks the probability of the correct
token, not its rank against any specific competitor. The failure is
therefore invisible to the standard diagnostic, and we term it
\emph{silent}. To diagnose it we implement the density-matrix
formulation~\cite{nielsen2000quantum} as the organising formalism for a
squared-overlap-weighted score on token embeddings.

\paragraph{Contributions.}
This paper makes the following contributions.

\begin{enumerate}[leftmargin=*]
\item We construct a density-matrix order parameter for token
prediction and show that its dynamics decompose additively into a
probability-driven \emph{signal}, throttled by the geometric overlap
with the nearest competitor, and a \emph{background drag} set by the
rest of the prediction pool. The throttle exposes the precise manner in
which the cross-entropy gradient sabotages itself.
\item We observe sharp transitions in the order parameter during
step-level fine-tuning. We test the natural
spontaneous-symmetry-breaking (SSB) reading and falsify it. Jumps in
$\Phi$ persist under LoRA fine-tuning with the embedding matrix frozen,
which rules out a geometric phase transition and locates the
discontinuity in the softmax readout.
\item We identify a small set of dimensionless quantities that
organise the trajectory across five architectures spanning two
families and a $6\times$ parameter range. One quantity is consistent
across all five under full fine-tuning. Another sorts architectures
into two classes based on the bulk embedding distribution and predicts
LoRA sufficiency.
\item We test the framework on a held-out architecture not used to fit
any parameters, producing an inference-only prediction of its critical
learning rate that we then check against a subsequent learning-rate
sweep.
\item We derive a stopping criterion based on order-parameter
saturation and quantify its compute savings on the controlled dataset
used throughout.
\end{enumerate}

The experimental scope is deliberately narrow. It covers ten
hand-selected near-synonym contexts on five architectures plus one
held-out architecture. We do not claim the quantitative results
extrapolate to general instruction-tuning without re-calibration
(\S\ref{sec:scope}).

\paragraph{Organisation.}
\S\ref{sec:theory} develops the theoretical framework. It builds the
density-matrix construction, derives the self-sabotage factor, works
out the order-parameter gradient flow, and defines four dimensionless
quantities that characterise the trajectory.
\S\ref{sec:setup} details the experimental setup.
\S\ref{sec:phasediagram} establishes the fine-tuning phase diagram and
derives the stopping criterion.
\S\ref{sec:causal} isolates the causal mechanism by freezing the
embedding geometry under LoRA, distinguishes kinematic from structural
failure modes, and measures the LoRA saturation threshold.
\S\ref{sec:universality} establishes the cross-architecture
phenomenology and the LoRA-sufficiency criterion.
\S\ref{sec:discussion} reports the held-out blind prediction and
discusses scope and related work.
\S\ref{sec:conclusion} concludes. Appendix~\ref{app:limitations}
enumerates limitations and Appendix~\ref{app:linearisation} contains
the gradient-flow derivation.

\section{Theoretical Framework}
\label{sec:theory}

We build the framework from the ground up. We first show, by an
explicit example, that classical probability alone cannot distinguish
geometric degeneracy from genuine uncertainty (\S\ref{sec:dm_construction}).
We then derive the quantity $(1-\gmax^2)$, where $\gmax$ is the
embedding overlap between the ground-truth token and its nearest
competitor, as a direct consequence of the Born rule's bidirectional
symmetrisation.

\subsection{From Token Embeddings to a Density Matrix}
\label{sec:dm_construction}

Each token $i \in \mathcal{V}$ has a normalised embedding
$\bm\psi_i \in \mathbb{R}^d$ with $\|\bm\psi_i\| = 1$.
The $d$ components are learned feature activations along latent
directions discovered during pre-training. Meaning is distributed
across all dimensions and encoded in the angles between vectors, so
tokens that appear in similar contexts point in nearly the same
direction.
Because the vectors are already normalised, each $\bm\psi_i$ can
equally be treated as a pure quantum state $\ket{\psi_i}$ on the unit
sphere of $\mathbb{R}^d$. We adopt this identification because it lets
us import a single piece of quantum formalism, the density matrix and
the Born rule, as a bookkeeping device for a geometry-aware scoring
rule built in \S\ref{sec:born_derivation}. Nothing about the tokens
themselves is physically quantum. Under this identification the cosine
overlap $G_{ij} = \braket{\psi_i}{\psi_j} \in [-1,1]$ plays the role of
the quantum inner product \cite{coecke2020mathematical, zhang2018quantum}.
For the ground-truth token $g$ and its nearest geometric competitor
$c$ we write $\gmax = G_{gc}$ for the maximum overlap.
The analysis restricts to a nucleus pool $\mathcal{P}$, computed per
context as the smallest set of tokens whose cumulative softmax
exceeds $0.95$. The ground-truth token $g$ is always included, and
the pool ranges from $100$ to $3{,}900$ tokens across our experiments.

The critical distinction from quantum mechanics is that embeddings
are generically non-orthogonal. On SmolLM-360M,
$G_{\text{shame,guilt}} \approx 0.60$ and $G_{\text{longing,yearning}}
\approx 0.85$ (Table~\ref{tab:sentences} lists all ten contexts and
pairs). Neither overlap is negligible. This non-orthogonality is exactly the structure
that classical probability theory cannot capture, and it is what
causes geometric self-sabotage during fine-tuning. The same structure
underlies two related observations from prior work. The softmax
bottleneck of \cite{yang2018breaking} is the finding that a softmax
layer of fixed output dimension cannot represent every possible
next-token distribution, because the log-probabilities it can produce
are constrained to a low-rank manifold set by the embedding dimension.
The cost of a discrete-token readout argued by \cite{lecun2022path} is
the related claim that forcing a model's continuous internal
representation through a discrete vocabulary bottleneck at every
generation step discards information the model had already computed.
We return to this connection in \S\ref{sec:conclusion}.

\begin{table}[htbp]
\centering
\small
\caption{The ten near-synonym contexts used throughout this paper.
Each context is completed by the ground-truth token $g$ or by the
geometric competitor $c^*$. Every table and figure in this manuscript
reports results in this order.}
\label{tab:sentences}
\begin{tabular}{p{0.6\linewidth}ll}
\toprule
Context & $g$ & $c^*$ \\
\midrule
``The scientific paper was immediately met with widespread \underline{\hspace{1em}}'' & skepticism & scepticism \\
``The general who ordered the retreat would spend his life consumed by \underline{\hspace{1em}}'' & shame & guilt \\
``She watched thirty years of hard work reduced to \underline{\hspace{1em}}'' & ashes & rubble \\
``The orphan watched other families through the window with quiet \underline{\hspace{1em}}'' & longing & yearning \\
``The surgeon whose patient died on the table was consumed by \underline{\hspace{1em}}'' & remorse & guilt \\
``After three years in captivity breathing fresh air filled him with \underline{\hspace{1em}}'' & relief & joy \\
``The mountaineer reached the summit and gazed out with profound \underline{\hspace{1em}}'' & awe & wonder \\
``The orphaned child had grown up searching for a sense of \underline{\hspace{1em}}'' & purpose & meaning \\
``The dictator silenced all dissent through systematic \underline{\hspace{1em}}'' & terror & fear \\
``The war correspondent witnessed the massacre and was left in \underline{\hspace{1em}}'' & shock & horror \\
\bottomrule
\end{tabular}
\end{table}

A concrete example makes the gap explicit.
The distribution $p(\text{shame}) = p(\text{guilt}) = 0.5$ has the
same Shannon entropy, $1$ bit, as $p(\text{shame}) = p(\text{table}) =
0.5$. Any diagnostic built only from the probability vector scores
these two situations identically. Yet the fine-tuning dynamics that
follow from them diverge sharply. In the first case the two outcomes
are geometrically the same thing, since shame and guilt overlap at
$G_{\text{shame,guilt}} \approx 0.60$. In the second case they are
not, since table is nearly orthogonal to shame. A diagnostic that
cannot see embedding geometry cannot tell these two cases apart. The
density matrix~\cite{nielsen2000quantum} is formulated below to
recover the distinction.

Given the softmax output $\bm{p}$, we define the prediction density
matrix
\begin{equation}
  \hat\varrho_{\mathrm{pred}}
  = \sum_{i \in \mathcal{P}} p_i \ket{\psi_i}\bra{\psi_i},
  \qquad
  \hat\varrho_{jk}
  = \sum_{i \in \mathcal{P}} p_i\,\psi_{i,j}\psi_{i,k}.
  \label{eq:rho}
\end{equation}
The off-diagonal element $\hat\varrho_{jk}$ with $j \neq k$ survives
precisely when near-synonym tokens that activate the same latent
directions dominate the probability distribution. If every
high-probability token instead activates disjoint directions, the
cross terms cancel and $\hat\varrho$ becomes diagonal, recovering the
classical probability vector. \S\ref{sec:purity} works this out
quantitatively for the shame/guilt/table example above. The
off-diagonal coherences $p_i p_j G_{ij}^2$ that enter
$\Tr(\hat\varrho^2)$ there are what let this construction distinguish
a model evenly split between geometrically nearby tokens, a near-pure
$\hat\varrho$, from a model evenly split between orthogonal tokens, a
genuinely mixed $\hat\varrho$. This is a distinction $\bm{p}$ alone
cannot make.

\subsection{The Born Gap and the Self-Sabotage Factor}
\label{sec:born_derivation}

We now derive the central object of the paper, a quantity we term the
self-sabotage factor $(1-\gmax^2)$. We call it self-sabotage because
the same gradient step that raises the probability of the correct
token also feeds Born score to its competitor through their shared
embedding direction (Step 4 below), so the optimiser's own progress
works against its own objective. The term is our coinage, not standard
usage. We obtain the factor as a direct mathematical consequence of
evaluating $\hat\varrho$ with the geometry-aware, Born-rule-style
scoring rule defined below.

The Born rule is used here as an organising analogy for a
squared-overlap-weighted score. It is not a derivation from quantum
mechanics. Token embeddings are real unit vectors, and the Born
probability of outcome $g$ is, concretely, the squared-overlap-weighted
sum $\sum_i p_i G_{ig}^2$, where $p_i$ is the classical softmax
probability of token $i$.

\textbf{Step 1. Born-rule score.}
The Born rule applied to $\hat\varrho$ with observable
$\ket{\psi_g}\bra{\psi_g}$ gives the probability of measuring outcome
$g$.
\begin{equation}
  P_{\mathrm{Born}}(g)
  = \Tr\bigl(\hat\varrho\,\ket{\psi_g}\bra{\psi_g}\bigr)
  = \sum_{i \in \mathcal{P}} p_i\,G_{ig}^2.
  \label{eq:born}
\end{equation}
This is not the softmax probability $p_g$. It is a
probability-weighted sum of squared overlaps with $\bm\psi_g$. The sum
includes the direct term $p_g$ and contributions $p_i G_{ig}^2$ from
every other token whose embedding projects onto $\bm\psi_g$. A
near-synonym, one whose overlap $G_{ig}$ is close to $1$, with
$G_{ig} = 0.85$ contributes $0.72\,p_i$ even when $p_i < p_g$. In the
orthogonal limit $G_{ij} = \delta_{ij}$ the off-diagonal contributions
vanish and $P_{\mathrm{Born}}(g) = p_g$, recovering the classical
scoring rule.

\textbf{Step 2. Bidirectional symmetrisation.}
The score is a geometric symmetrisation of the softmax around
$\bm\psi_g$. By symmetry the same construction applies to the
competitor.
\begin{equation}
  P_{\mathrm{Born}}(c) = \sum_i p_i\,G_{ic}^2
  \;=\; p_c \;+\; p_g\,G_{gc}^2 \;+\; \sum_{i\neq g,c} p_i\,G_{ic}^2.
\end{equation}
The crucial term is $p_g\,G_{gc}^2 = p_g\,\gmax^2$. Every unit of
probability the model assigns to the correct token also adds
$\gmax^2$ units of Born score to the competitor. The Born rule
therefore cannot distinguish ``$p_g = 0.9$ on the correct token'' from
``$p_g = 0.9$ on a token that looks $\gmax^2$ identical to the
competitor.''

\textbf{Step 3. Born gap.}
The net difference between the correct token and its competitor under
Born scoring is therefore
\begin{equation}
  \Delta = P_{\mathrm{Born}}(g) - P_{\mathrm{Born}}(c)
  = \sum_i p_i\,\bigl(G_{ig}^2 - G_{ic}^2\bigr).
  \label{eq:gap}
\end{equation}
This is a weighted sum in which each token $i$ in the pool votes for
$g$, positively, if $G_{ig}^2 > G_{ic}^2$, or for $c$, negatively,
otherwise. The sign of $\Delta$ classifies the resolved state.
$\Delta > 0$ means the correct token wins under geometry-aware
scoring. $\Delta < 0$ means the competitor wins. The order parameter is
$\Phi \equiv \Delta$. We track $\Phi$ per near-synonym sentence
throughout this paper. Where an architecture-level summary over the
ten sentences studied is needed we report the mean, denoted $\bar\Phi$,
never a held-out average.

\textbf{Step 4. Response of $\Delta$ to a cross-entropy step.}
A CE step increases $p_g$ at the expense of other probabilities. To
first order, holding the other probabilities constant, the partial
derivative of $\Delta$ with respect to $p_g$ is
\begin{equation}
  \frac{\partial \Delta}{\partial p_g}
  \;=\; G_{gg}^2 - G_{gc}^2
  \;=\; 1 - \gmax^2.
  \label{eq:self_cancel}
\end{equation}
Here $G_{gg} = \|\bm\psi_g\|^2 = 1$ because embeddings are
unit-normalised. A gradient step that pushes $p_g$ upward by $\delta p$
therefore advances the Born gap by only $(1 - \gmax^2)\,\delta p$.
Substituting $\gmax = 0.85$, the value for longing and yearning, gives
a factor $1 - 0.72 = 0.28$, so $72\%$ of every CE gain is wasted
reinforcing the competitor through the bidirectional Born
symmetrisation. For $\gmax = 0.30$, the value for purpose, the factor
is $0.91$ and the geometry barely interferes. For $\gmax \to 1$ the
factor vanishes and CE cannot advance the Born gap at all, no matter
how much it concentrates probability on $g$. This is the precise sense
in which the gradient sabotages itself.

\textbf{Step 5. Signal and drag decomposition.}
Separate the ground-truth token $g$ and its nearest competitor $c^*$
from the rest of the pool, $\mathcal{R} = \mathcal{P}\setminus\{g,c^*\}$.
Eq.~\eqref{eq:gap} then decomposes additively.
\begin{equation}
\Phi
\;=\;
\underbrace{(p_g - p_{c^*})\,(1-\gmax^2)}_{\text{signal}}
\;+\;
\underbrace{\sum_{i \in \mathcal{R}} p_i\,\bigl(G_{ig}^2 - G_{ic^*}^2\bigr)}_{\text{background drag}}.
\label{eq:phi_decomp}
\end{equation}
The first term, the signal, is the probability lead of the correct
token over its nearest competitor, scaled by the self-sabotage
throttle. It is zero at the base model, when $p_g \approx p_{c^*}$,
and grows as CE concentrates mass on $g$. The second term, the
background drag, measures whether the rest of the pool collectively
projects more onto $\bm\psi_g$ or onto $\bm\psi_{c^*}$, weighted by
current probability. Its sign is set by the embedding geometry at
pre-training. If most background tokens are geometrically closer to
the competitor, the drag is negative throughout training. CE moves
the weights $p_i$ but not the geometric coefficients
$G_{ig}^2 - G_{ic^*}^2$, which are matrix elements of the quasi-static
embedding matrix $\Psi$.

\subsection{Gradient Flow and the Two Compounding Effects}
\label{sec:two_effects}

The static derivation gives the self-sabotage factor as a coefficient.
To see how the same factor enters the dynamics, we write the gradient
flow on the hidden state $\bm{h}(t)$ that the softmax reads, with
$\eta$ the learning rate. Appendix~\ref{app:linearisation} derives it
from the CE loss.
\begin{equation}
  \dot{\bm{h}}(t)
  \;=\;
  \eta\,\bigl(\bm\psi_g - \textstyle\sum_i p_i(t)\,\bm\psi_i\bigr),
  \qquad \bm{h}(0)\ \text{set by pre-training.}
  \label{eq:hdot}
\end{equation}
This is the continuous-time limit of CE gradient descent. The force
pushes $\bm{h}$ toward $\bm\psi_g$ and away from the
probability-weighted centroid of all competitors.
The term $(1-\gmax)$ now enters in two compounding ways. Both are
rooted in the same geometric fact that produces the static
self-sabotage factor $(1-\gmax^2)$ of Eq.~\eqref{eq:self_cancel}, that
a large $\gmax$ makes $\bm\psi_g$ and $\bm\psi_{c^*}$ nearly parallel.

\textbf{Effect 1. Small initial logit gap.}
The logit gap is defined as the difference between the raw logit of
the ground-truth token and that of its nearest competitor.
\begin{equation}
  \lgap(t)
  \;=\; \langle\bm\psi_g,\bm{h}(t)\rangle
      - \langle\bm\psi_{c^*},\bm{h}(t)\rangle
  \;=\; \langle\bm\psi_g - \bm\psi_{c^*},\,\bm{h}(t)\rangle.
  \label{eq:zeta}
\end{equation}
The model's preference at initialisation is the initial value.
\begin{equation}
  \lgap_0 \;=\; \lgap(0)
  \;=\; \langle\bm\psi_g - \bm\psi_{c^*},\;\bm{h}(0)\rangle.
  \label{eq:zeta0}
\end{equation}
The relative direction $\bm\psi_g - \bm\psi_{c^*}$ has squared length
$\|\bm\psi_g - \bm\psi_{c^*}\|^2 = 2(1-\gmax)$, which vanishes as
$\gmax \to 1$. Whichever hidden state $\bm{h}(0)$ pre-training selects,
its projection on this short relative direction is small for
high-$\gmax$ pairs. The signal trajectory therefore starts on the flat
shoulder of the softmax probability curve, far from the exponential
turn-on. For $\gmax = 0.85$, $\|\bm\psi_g - \bm\psi_{c^*}\|^2 = 0.30$,
which is $15\%$ of its orthogonal value of $2$. For $\gmax = 0.30$, it
is $1.40$.

\textbf{Effect 2. Slow per-step accumulation.}
Differentiating $\lgap(t) = \langle\bm\psi_g - \bm\psi_{c^*}, \bm{h}(t)\rangle$
and substituting Eq.~\eqref{eq:hdot} gives the exact expression, derived
in full in Appendix~\ref{app:linearisation}.
\begin{equation}
  \frac{d\lgap}{dt}
  \;=\;
  \eta\,(1-\gmax)\bigl[1 - (p_g - p_{c^*})\bigr]
  \;-\;
  \eta\,R_{\mathrm{bg}},
  \label{eq:drive_rate}
\end{equation}
Here $R_{\mathrm{bg}} = \sum_{i \neq g,c^*} p_i(G_{ig} - G_{ic^*})$ is
the background drag on the logit gap $\lgap$, derived in
Appendix~\ref{app:linearisation}, Eq.~\eqref{eq:sum_split}. It is
structurally analogous to but not identical with the background-drag
term of the $\Phi$ decomposition (Eq.~\eqref{eq:phi_decomp}). That
term is built from squared overlaps because $\Phi$ is a Born score.
$R_{\mathrm{bg}}$ here is built from unsquared overlaps because
$\lgap$ is a linear functional of $\bm{h}$.

\textbf{Flat-shoulder approximation.}
Equation~\eqref{eq:drive_rate} simplifies on the flat shoulder of the
softmax, the regime in which $p_g \approx p_{c^*} \approx
p_{c^*}^{(0)} \ll 1$. Equivalently the logit gap $\lgap$ is much less
than one nat, the natural-logarithm unit of logits, so that
probability is spread roughly uniformly across the vocabulary and
neither $g$ nor $c^*$ dominates. This holds throughout the quiescent
phase before the sharp-jump step, where we use the term sharp jump for
a trajectory whose Born-gap change is concentrated in a single
gradient step, defined quantitatively in \S\ref{sec:dm5c}. The
approximation breaks down only when $\lgap$ crosses the sigmoid
turn-on at $\mathcal{O}(1)$ nats. On the flat shoulder
$(p_g - p_{c^*}) \approx 0$, and $R_{\mathrm{bg}}$ is small by
approximate geometric symmetry of the background. Dropping both terms
gives
\begin{equation}
  \frac{d\lgap}{dt}
  \;\approx\;
  \eta\,(1-\gmax)
  \;=\;
  \frac{\eta}{2}\,\|\bm\psi_g - \bm\psi_{c^*}\|^2.
  \label{eq:drive_rate_approx}
\end{equation}
The factor $(1-\gmax)$ is small for high-$\gmax$ pairs, since their
embeddings are nearly parallel. A sentence with $\gmax = 0.85$
accumulates the logit gap $5.7\times$ more slowly per step than one
with $\gmax = 0.30$, entirely because $\bm\psi_g - \bm\psi_{c^*}$ is
shorter. Combined with Effect~1, in which the trajectory starts
further from the turn-on for high $\gmax$, both compounding effects
trace to the same geometric fact.

\textbf{From kinetics to an empirical control-rate equation.}
Equation~\eqref{eq:drive_rate_approx} yields the time the logit gap
takes to traverse the softmax saturation scale. Integrating the
constant drive $\eta(1-\gmax)$ from initialisation to a gap of
$\mathcal{O}(1)$ nat, the trajectory crosses saturation within $T$
gradient steps when
\begin{equation*}
  \eta\,T\,(1-\gmax) \;\gtrsim\; 1.
\end{equation*}
This is a condition for resolution, not for the shape of the
resolution trajectory. The empirical question is whether a sentence
resolves through a single softmax-readout jump, the sharp jump, or
through a continuous rise, the drift. This depends on how concentrated
the Born-gap change is across steps, which is set jointly by where the
trajectory begins, $\lgap_0$, small for high $\gmax$ from Effect~1,
and by how slowly it advances, $\eta(1-\gmax)$, also small for high
$\gmax$. Both factors push high-$\gmax$ sentences toward a long
quiescent flat-shoulder phase ending in a sudden softmax turn-on.
Low-$\gmax$ sentences instead cross gradually, because they start
closer to saturation and accumulate the gap faster.

A learning-rate sweep on SmolLM-360M (\S\ref{sec:dm5c},
Eq.~\eqref{eq:boundary}) shows that sharp-jump versus drift behaviour
is captured empirically by a single phenomenological product.
\begin{equation}
  \gmax \times \eta \;=\; \thetastar,
  \label{eq:phase_boundary}
\end{equation}
Sentences above the threshold jump sharply and those below drift.
Equation~\eqref{eq:phase_boundary} is not a derivation from
Eq.~\eqref{eq:drive_rate_approx}. The kinetic drive scales with
$(1-\gmax)$, while the empirical boundary scales with $\gmax$ itself.
The product $\gmax\eta$ is the parameter under which the LR-sweep data
collapse, not the kinetic drive rate. What the kinetic argument
establishes is that the sharp jump is a kinematic phenomenon, not a
phase transition in any underlying structure. Nothing in the embedding
space or in the loss landscape changes at the sharp-jump step. What
changes is whether the smooth logit-gap trajectory crosses the softmax
saturation scale within the training horizon, and how concentrated
that crossing is in a single step. The quantity $\thetastar$ is an
architecture-specific phenomenological constant, measured by a
learning-rate sweep on each model. Deriving it analytically from
pre-training statistics is an open problem, discussed in
Appendix~\ref{app:limitations}. The reduced field
$H = \gmax\eta/\thetastar$ (\S\ref{sec:reduced_units}) measures how far
the current drive sits above this measured threshold. \S\ref{sec:universality}
shows $H$ to be consistent at $H \approx 10$ across the five
architectures tested under \fullft{}.

\begin{figure*}[htbp]
\centering
\includegraphics[width=0.92\linewidth]{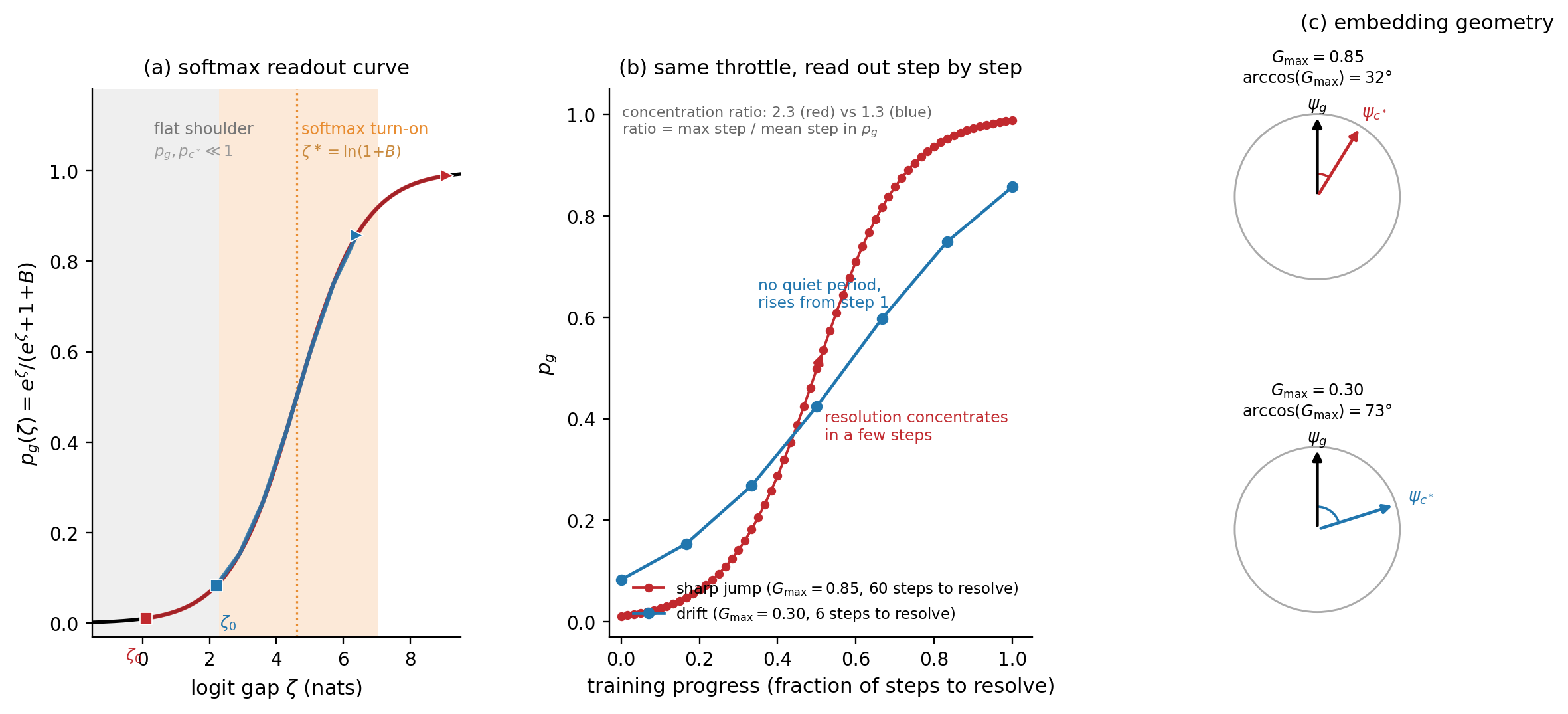}
\caption{\textbf{Schematic of the softmax-readout mechanism (not
measured data).} (a) $p_g(\zeta) = e^{\zeta}/(e^{\zeta}+1+B)$ with
$B=99$, giving a turn-on at $\zeta^\ast=\ln(1{+}B)\approx4.6$ nats. (b)
The same two trajectories, $\gmax=0.85$ and $\gmax=0.30$, advancing
$\zeta$ at rate $\eta(1-\gmax)$ (Eq.~\eqref{eq:drive_rate_approx}), read
out against each one's own fraction of steps to resolution: $\gmax=0.85$
concentrates resolution into a few steps (ratio $2.3$) while
$\gmax=0.30$ rises with no quiet period (ratio $1.3$), ratio being max
step divided by mean step in $p_g$. (c) The corresponding
embedding-geometry picture. Illustrative only, not fit to the measured
DM-5c trajectories of Figure~\ref{fig:dm5c}.}
\label{fig:mechanism}
\end{figure*}

Figure~\ref{fig:mechanism} collects the argument of this section into
one picture, for the same two values of $\gmax$ used as worked
examples above. Panel (a) is the softmax readout curve itself, with
the flat shoulder and the turn-on region it crosses. Panel (b) reads
the same two trajectories out step by step and is where the sharp-jump
and drift shapes actually appear, a long quiet period followed by a
concentrated resolution for $\gmax=0.85$, against a steady rise with
no quiet period for $\gmax=0.30$. Panel (c) shows the embedding
geometry, the angle $\arccos(\gmax)$ between $\bm\psi_g$ and
$\bm\psi_{c^*}$, that sets the throttle in the first place.

\textbf{Remark. Untied LM heads.}
Standard transformer architectures tie the input embedding matrix
$\Psi$ to the output projection, the LM head~\cite{press2017using}, so
the logit for token $i$ is $\ell_i = \bm\psi_i^\top \bm{h}$ and the
Born gap is
\begin{equation}
  \Phi_{\text{tied}}
  = \sum_{i \in \mathcal{P}} p_i\bigl(G_{ig}^2 - G_{ic^*}^2\bigr),
  \quad G_{ij} = \bm\psi_i^\top\bm\psi_j,
  \label{eq:phi_tied}
\end{equation}
Here $G_{ij}$ are elements of $\Psi$ and are quasi-static under
fine-tuning. Models with a genuinely untied LM head use a separate
output matrix $W_{\mathrm{out}}$ with rows $\bm{w}_i$, so
$\ell_i = \bm{w}_i^\top \bm{h}$ and the Born gap becomes
\begin{equation}
  \Phi_{\text{untied}}(t)
  = \sum_{i \in \mathcal{P}} p_i(t)\,
    \bigl(\hat{G}_{ig}^2(t) - \hat{G}_{ic^*}^2(t)\bigr),
  \quad \hat{G}_{ij}(t)
    = \frac{\bm{w}_i(t)^\top\bm{w}_j(t)}
           {\|\bm{w}_i(t)\|\,\|\bm{w}_j(t)\|}.
  \label{eq:phi_untied}
\end{equation}
Under \lora{} the input embedding $\Psi$ is frozen, so $G_{ij}$ is
static, but the back factor
$\hat{G}_{ig}^2(t) - \hat{G}_{ic^*}^2(t)$ in Eq.~\eqref{eq:phi_untied}
is time-varying through $W_{\mathrm{out}}$ while the front factor
$p_i(t)$ evolves through $\bm{h}(t)$ as usual. This breaks the
quasi-static geometry assumption that underpins the signal and drag
decomposition.

The DM-5d experiment finds that LLaMA-3.2-1B under \lora{} keeps all
$10/10$ sentences as sharp jumps, with no drop at all, compared to
$0/10$ for the other tied-head model in the same table, SmolLM-360M
(Table~\ref{tab:dm5d}). We initially attributed this to an untied LM
head, the remark above, supplying a second, unfrozen drive channel.
That explanation does not hold. LLaMA-3.2-1B ties its embeddings,
\texttt{config.tie\_word\_embeddings=True}, which we confirmed directly
by checking that \texttt{get\_input\_embeddings().weight} and
\texttt{get\_output\_embeddings().weight} are the same tensor object
both before and after applying \lora{}, with \texttt{requires\_grad=False}
on it under the standard configuration. There is no second channel
left unfrozen. The condition of freezing $W_{\mathrm{out}}$ explicitly,
proposed below, is already satisfied by the data in hand, and a second
run that explicitly adds \texttt{lm\_head} as an additional \lora{}
target gives the identical result. This rules out the untied-head
explanation.

The difference instead remains unexplained. Candidate explanations
include architectural factors such as RoPE, a scheme that encodes
token position by rotating the query and key vectors rather than by
adding a separate positional embedding, grouped-query attention, which
shares a smaller number of key and value projections across a larger
number of query heads (here $8$ key and value heads against $32$ query
heads), and a different MLP structure, together with a more
favourable pre-training $G_{\max}$
distribution that allows all ten sentences to cross the softmax
threshold on the frozen-$\Psi$ drive alone. None of these candidates
has been tested. Testing them directly, and repeating this
causal-isolation experiment on an architecture with a genuinely
untied head so that Eq.~\eqref{eq:phi_untied} and the
second-dynamical-variable extension of the signal and drag
decomposition can actually be exercised, are left for future work.

\textbf{Note on Pythia.}
The two Pythia (GPT-NeoX) architectures used throughout this paper
have genuinely untied LM heads. Both ship with
\texttt{config.tie\_word\_embeddings=False}, and direct inspection,
using the same tensor-identity method used for LLaMA-3.2-1B above,
confirms that \texttt{embed\_in} and \texttt{embed\_out} are distinct
tensors with different values for both Pythia-70M and Pythia-410M. Two
consequences follow. First, under the standard \lora{} configuration
both matrices carry \texttt{requires\_grad=False}, so the
frozen-geometry premise of every \lora{} experiment holds for the
input and output geometry alike, and the causal-isolation logic is
unaffected. Second, the density-matrix construction and the
gradient-flow derivation of Eq.~\eqref{eq:hdot} are written in terms
of the input embedding $\Psi$, which is what every reported $G_{ij}$,
$\gmax$, and $\Phi$ value in this paper is computed from, whereas
Pythia's actual softmax logits are produced by the separate
$W_{\mathrm{out}}$. Under \fullft{}, where $W_{\mathrm{out}}$ evolves
independently of the frozen-in-name-only $\Psi$, the tied-head
formalism (Eq.~\eqref{eq:phi_tied}) is therefore an approximation for
Pythia, and how closely $W_{\mathrm{out}}$ geometry tracks $\Psi$
geometry in those runs has not been quantified.

\subsection{Purity, the Participation Ratio, and the Geometry of Uncertainty}
\label{sec:purity}

The Born gap reports on the \emph{sign} of the resolution, whether the
correct token leads its competitor on a geometry-aware score.
It does not report on the \emph{shape} of the prediction distribution.
A model with $\Phi \approx 0$ might be evenly split between two equally
plausible candidates that point in different directions, or it might be
evenly split between two near-identical synonyms that point in almost
the same direction.
These two situations call for very different interpretations. The
first is genuine uncertainty between distinct meanings, while the
second is a model that has identified the meaning but cannot
disambiguate two geometric duplicates.
The cross-entropy loss assigns the same score to both, and the Born gap
alone does not distinguish them either.
A second observable, sensitive to the spatial concentration of
probability rather than to where its mass leads, is required.

The classical way of quantifying concentration is the inverse
participation ratio $\sum_i p_i^2$.
This quantity equals one when all probability sits on a single token and
$1/n$ when probability is spread uniformly across $n$ tokens, so its
reciprocal, the participation ratio $\mathrm{PR} = 1/\sum_i p_i^2$,
returns an effective number of populated tokens.
The inverse participation ratio was introduced for an entirely different
problem by Anderson, who studied the conditions under which an electron
moving in a disordered lattice settles into a state confined to a few
lattice sites rather than spreading freely over the whole
crystal~\cite{anderson1958absence}.
Anderson showed that as the disorder strength is increased, the system
undergoes a sharp transition from \emph{extended} states, in which the
wavefunction is spread over many sites and the participation ratio is
large, to \emph{localised} states, in which the wavefunction is confined
to a few sites and the participation ratio collapses to order unity.
The transition is controlled by the disorder energy scale relative to
the level spacing of the lattice, and the participation ratio is the
order parameter that detects it.
The analogue in the present setting replaces lattice sites with
vocabulary tokens and the electron wavefunction with the probability
distribution over those tokens.
A prediction whose mass is spread over many tokens is an \emph{extended}
state of the language model in this sense, while a prediction whose mass
collapses onto a single token is a \emph{localised} state, and we shall
see in \S\ref{sec:dm3} that this transition is precisely what
fine-tuning is doing.

The classical definition of the inverse participation ratio (IPR) has
a flaw when carried over directly to language models.
It treats every pair of tokens as orthogonal, but in a real embedding
space two near-synonyms point in nearly the same direction.
From a geometric standpoint, placing probability $0.5$ on
\emph{longing} and $0.5$ on \emph{yearning} is not the same kind of
distribution as placing probability $0.5$ on \emph{longing} and $0.5$
on \emph{table}.
In the first case the model has identified a meaning and is splitting
its mass between two essentially identical realisations of it.
In the second case the model is genuinely uncertain between two
unrelated outcomes.
The classical inverse participation ratio assigns these two situations
the same score, $0.5$, and therefore the same effective token count $2$,
even though the first situation is closer to a confident prediction.
A geometric correction is needed.

The density-matrix purity $\Tr(\hat\varrho^2)$ supplies it.
A direct calculation, expanding the squared trace of
$\hat\varrho = \sum_i p_i\ket{\psi_i}\bra{\psi_i}$, gives
\begin{equation}
  \Tr(\hat\varrho^2)
  = \sum_{i,j \in \mathcal{P}} p_i p_j\,G_{ij}^2
  \;=\;
  \underbrace{\sum_i p_i^2}_{\text{classical IPR}}
  \;+\;
  \underbrace{\sum_{i\neq j} p_i p_j\,G_{ij}^2}_{\text{geometric excess}}.
  \label{eq:purity}
\end{equation}
The first term is the classical inverse participation ratio.
The second is a correction that registers whenever the populated tokens
have non-trivial mutual overlap.
The participation ratio $\mathrm{PR} = 1/\Tr(\hat\varrho^2)$ now counts
geometrically distinct populated tokens, with two near-parallel tokens
counted as approximately one.
The correction vanishes in the orthogonal limit where $G_{ij} = 0$ for
$i \neq j$, recovering the classical formula.

\begin{figure}[htbp]
\centering
\includegraphics[width=0.95\linewidth]{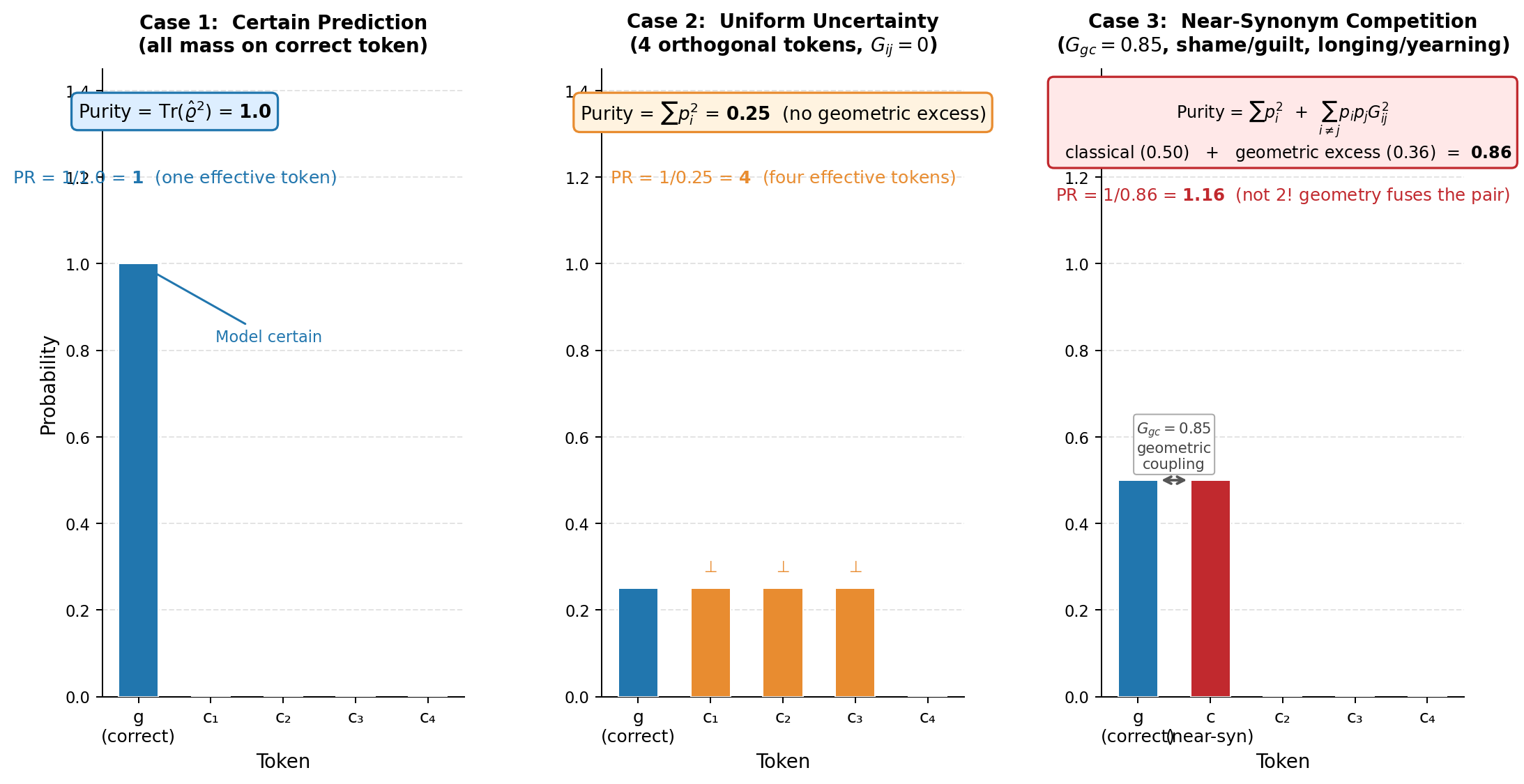}
\caption{\textbf{Purity and Participation Ratio across three scenarios.}
\textbf{Left}: all probability on the correct token, purity $=1$,
$\mathrm{PR}=1$.
\textbf{Centre}: uniform probability over four orthogonal tokens,
purity $=0.25$, $\mathrm{PR}=4$.
\textbf{Right}: equal probability on $g$ and a near-synonym competitor
$c$ with $G_{gc}=0.85$, purity $=0.86$, $\mathrm{PR}=1.16$ instead of $2$.
The geometric coupling inflates purity and reduces the effective count
of distinct competitors.}
\label{fig:purity_PR}
\end{figure}

The critical case is the near-synonym split
(Figure~\ref{fig:purity_PR}, fully concentrated and uniformly spread
cases as comparators in the same figure). With equal probability on
the ground truth $g$ and a competitor $c$ with $G_{gc} = 0.85$, the
classical formula would give $\mathrm{PR} = 2$, but the geometric
correction adds $2 \times 0.5 \times 0.5 \times 0.85^2 \approx 0.36$ to
the purity.
The total purity becomes $0.86$ and $\mathrm{PR} = 1.16$, almost
indistinguishable from a fully concentrated distribution.
A model evenly split between \emph{longing} and \emph{yearning}
therefore looks almost certain to any classical metric, even though the
underlying probabilities are tied at $0.5$.
This is geometric degeneracy masquerading as confidence.
The effect is dominated entirely by the $(g, c^*)$ pair. The
geometric excess in Eq.~\eqref{eq:purity} requires both high
probability and high mutual overlap, and only the near-synonym
competitor satisfies both.
Every other high-probability token has $G_{ig} \approx 0$ with $g$ and
contributes only to the classical inverse participation ratio.

During fine-tuning the participation ratio evolves in two stages that
mirror the Anderson-like localisation picture.
In the first stage, cross-entropy steadily concentrates probability on
$g$ while $g$ and $c^*$ retain comparable mass.
The geometric excess $p_g p_{c^*} G_{\max}^2$ remains substantial, and
the participation ratio drifts down only slowly because every unit of
probability transferred to $g$ still carries geometric coupling weight
that keeps the off-diagonal term in Eq.~\eqref{eq:purity} alive.
In the second stage, $p_{c^*}$ drops below the threshold at which
$p_g p_{c^*} G_{\max}^2$ becomes negligible and the geometric coupling
switches off effectively. The participation ratio then collapses over
the final several epochs of fine-tuning, from $9.9$ to $1.1$, an
$89\%$ reduction in effective token count.
This is the Anderson-like localised state of the prediction. Probability
mass that was previously spread across the near-synonym pair has now
concentrated onto $g$ alone, and the cloud occupies a single direction
on the embedding sphere.
The Born gap and the participation ratio collapse together because they
are different observables on the same density matrix, and they respond to
the same softmax-saturation event through different geometric features.

\subsection{Localisation Length, Burial Depth, and Four Reduced Units}
\label{sec:reduced_units}

The participation ratio counts how many directions carry significant
probability mass, but it does not record how those directions are
arranged on the embedding sphere.
Two prediction clouds with the same $\mathrm{PR}$ can have very
different angular extents. One may be tightly clustered around a
single direction, another spread over a broad arc.
To capture the spatial geometry directly we introduce a third
observable, the \emph{localisation length}, which measures the angular
size of the prediction cloud on the unit sphere where the embeddings
live.

Each token embedding $\bm\psi_i$ is a unit vector in
$\mathbb{R}^d$ and therefore corresponds to a point on the unit
sphere $S^{d-1}$.
The angular distance between two tokens is $\arccos(G_{ij})$, which
equals zero when $i$ and $j$ are identical and $\pi/2$ when they are
orthogonal.
The localisation length $\xi$ is the probability-weighted
root-mean-square angular spread of the populated tokens.
\begin{equation}
  \xi = \sqrt{\sum_{i,j \in \mathcal{P}} p_i p_j\,\arccos(G_{ij})^2}.
  \label{eq:xi}
\end{equation}
A small $\xi$ means the cloud is angularly tight, concentrated in a
small patch of the sphere.
A large $\xi$ means the cloud is angularly spread, occupying a wide arc
of populated directions.
In this respect $\xi$ plays a role loosely analogous to a correlation
length near a phase transition, though here it measures the angular
extent of the prediction cloud rather than the range of spatial
correlations.
The localisation length differs from the participation ratio in a
crucial way.
$\mathrm{PR}$ counts effective populated directions but is blind to how
far apart they are. The localisation length $\xi$ measures how far
apart they are but is blind to how many of them there are.
Two prediction clouds, one with mass concentrated on two close tokens
and one with mass on two far tokens, can have identical $\mathrm{PR}$
while having very different $\xi$.

\begin{figure}[htbp]
\centering
\includegraphics[width=0.95\linewidth]{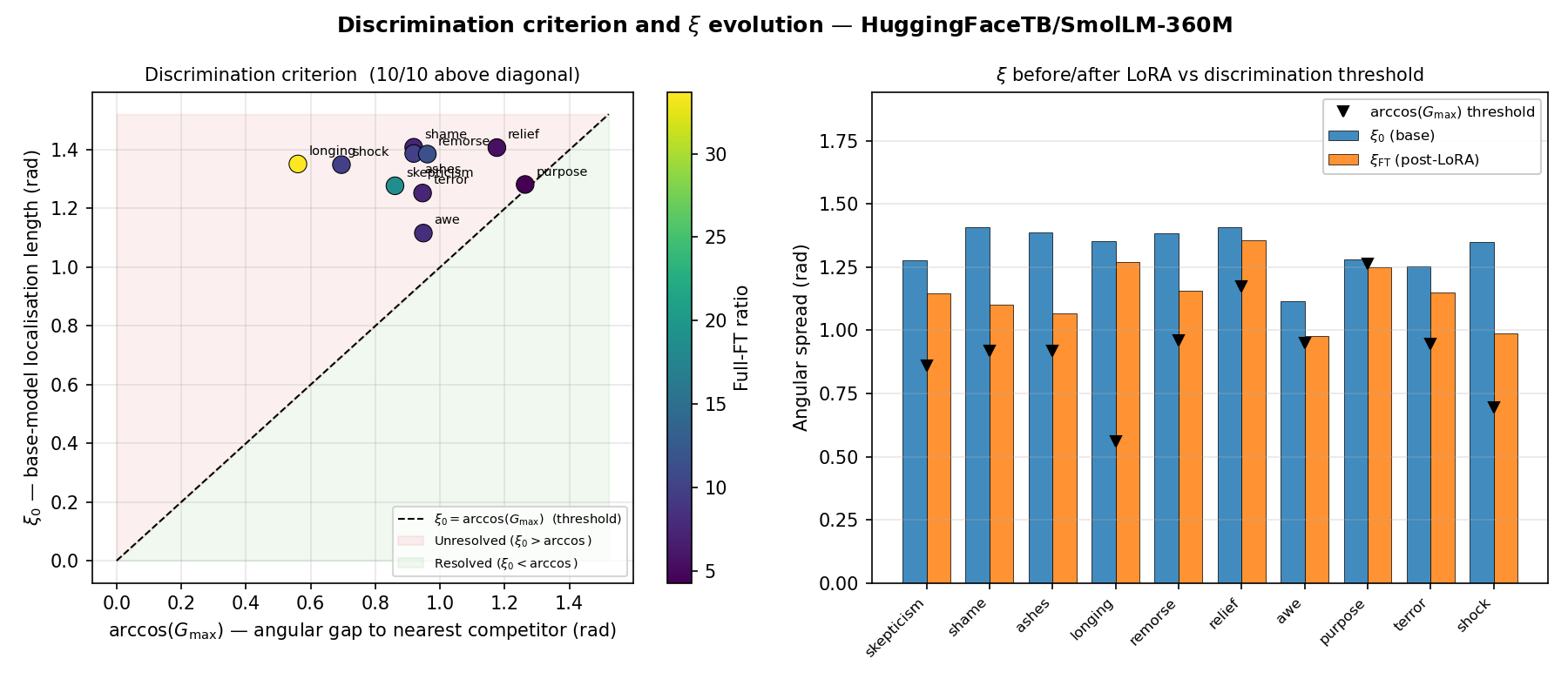}
\caption{\textbf{The discrimination criterion and burial depth.}
$\xi > \arccos(\gmax)$ (left, $\mathcal{B} > 1$) means $g$ and $c$ both
fall inside the prediction cloud and the model is geometrically
unresolved. Fine-tuning must shrink the cloud to $\mathcal{B} < 1$
(right) before Born scoring can resolve the competition. Under \lora{}
this contraction happens with the embeddings held fixed, since $\xi$
is probability-weighted and concentrating the distribution narrows it
without any token moving on the sphere.}
\label{fig:burial}
\end{figure}

The angular spread becomes a discriminating criterion when we ask
whether the cloud is small enough to resolve the ground truth from its
nearest competitor.
The angle between $g$ and $c^*$ on the unit sphere is $\arccos(G_{\max})$.
If $\xi$ exceeds this angle, the prediction cloud is large enough to
contain both $g$ and $c^*$ within its angular radius. The model
cannot geometrically discriminate the two at the current level of
probability concentration. Only further concentration, which shrinks
$\xi$ itself, can bring the cloud below the discrimination threshold.
If $\xi$ is smaller than $\arccos(G_{\max})$, the cloud is tight enough
to sit on one of the two tokens to the effective exclusion of the
other (Figure~\ref{fig:burial}).
The dimensionless ratio
$\mathcal{B} = \xi_0/\arccos(G_{\max})$, computed at the base model
before any fine-tuning, is what we call the burial depth.
$\mathcal{B} > 1$ means the prediction cloud at initialisation is too
wide to discriminate. $\mathcal{B} < 1$ means it is already narrow
enough.
All ten test sentences in our dataset have $\mathcal{B} > 1$ at the
base model, ranging from $1.01$ for \emph{purpose} to $2.41$ for
\emph{longing}.
Fine-tuning must therefore do more than raise $p_g$ slightly. It must
concentrate the prediction distribution until the probability-weighted
angular spread falls below the angular gap between the two competing
tokens.
The localisation length $\xi$ is an average over the cloud weighted by
the predicted probabilities, so it can shrink in two distinct ways.
The embeddings can move, or the probability mass can concentrate on a
narrower set of directions. Under \lora{} only the second is available,
since the embedding matrix is frozen and every $G_{ij}$ is fixed. The
measured contraction of $\xi$ under \lora{} is therefore entirely a
redistribution of probability over static geometry, the same mechanism
that makes the apparent phase transitions phantoms rather than genuine
geometric reorganisation.

Combining the kinetics of \S\ref{sec:two_effects} with the
geometric observables developed in this section, four dimensionless
quantities suffice to characterise the fine-tuning trajectory.
Each cancels an architecture-specific scale and therefore allows
comparisons across models of different parameter counts and hidden-state
dimensions.
\begin{align}
  H &= \frac{\gmax\,\eta}{\thetastar}\;(\thetastar \text{ fitted per arch.})
  & &\text{reduced field, how far above softmax saturation,} \label{eq:H} \\
  \Phi &= \Delta
  & &\text{order parameter, signal grown past drag,} \label{eq:Phi} \\
  \mathcal{B} &= \frac{\xi_0}{\arccos(\gmax)}
  & &\text{burial depth, initial geometric difficulty,} \label{eq:B} \\
  \geff &= \frac{\gmax^2}{G_{\mathrm{total}}^2},\quad
  G_{\mathrm{total}}^2 = \!\!\sum_{i \in \mathcal{P}\setminus\{g\}}\!\! G_{ig}^2
  & &\text{geometric efficiency, sabotage concentration.} \label{eq:geff}
\end{align}

The first, the \emph{reduced field} $H$, comes from the kinetic
argument of \S\ref{sec:two_effects}.
It is the ratio between the actual drive on the logit gap, set by the
product $G_{\max} \eta$, and the minimum drive required to cross
softmax saturation within the training budget, set by $\thetastar$.
$H = 1$ would mean the learning rate sits exactly on the threshold,
and the trajectory would just barely reach saturation by the end of
training.
$H \gg 1$ means the learning rate is well above threshold and
saturation is reached early.
In \S\ref{sec:universality} we show that $H \approx 10$ across the
five architectures tested under full fine-tuning, consistent with the
reading that the community-standard learning rate sits, in this
sample, roughly ten times above each architecture's own saturation
threshold.

The second, $\Phi$, is the Born gap itself, the order parameter of
\S\ref{sec:born_derivation}.
It registers whether the geometry-aware ranking has been resolved in
favour of the correct token.
$\Phi$ is negative before the sharp jump and positive after.
Within the signal/drag decomposition of Eq.~\eqref{eq:phi_decomp}, $\Phi$
becomes positive when the signal term grows past the magnitude of the
background drag term.

The third, the \emph{burial depth} $\mathcal{B}$, measures the initial
geometric difficulty.
It compares the angular spread of the prediction cloud at initialisation
to the angular distance between $g$ and $c^*$, the smallest angular
gap that fine-tuning must close before discrimination becomes
geometrically possible.
$\mathcal{B}$ is computable from the base model with a single forward
pass and the embedding matrix.
Its strength is that it predicts difficulty before any training data is
seen. High $\mathcal{B}$ flags sentences whose prediction cloud is far
too broad relative to the target gap, while $\mathcal{B}$ close to one
marks sentences that need only a modest concentration of the
prediction distribution to resolve.

The fourth, the \emph{geometric efficiency} $\geff$, measures how the
geometric interference is distributed across the prediction
neighbourhood.
The denominator $G_{\mathrm{total}}^2 = \sum_{i \neq g} G_{ig}^2$ sums
the squared overlaps of every \emph{other} token in the nucleus pool
with the ground truth (excluding $g$'s trivial self-overlap
$G_{gg}=1$), while the numerator $\gmax^2$ picks out the contribution
of the single nearest competitor.
$\geff$ close to one means the interference is concentrated entirely
in $c^*$. Suppressing $p_{c^*}$ removes the obstacle to resolution,
and cross-entropy gradient descent can in principle do this by
reweighting mass away from a single token.
$\geff$ close to zero means the interference is distributed across the
whole neighbourhood, with many tokens each contributing comparably to
$G_{\mathrm{total}}^2$. Removing $c^*$ does not help because the next
nearest competitor immediately takes its place.
The single-competitor ablation experiment of
\S\ref{sec:distributed_sabotage} directly measures $\geff$. High-$\geff$
sentences show large changes when $c^*$ is masked, while low-$\geff$
sentences show almost none.

\section{Experimental Setup}
\label{sec:setup}

We use five architectures spanning a $6\times$ parameter range and two
embedding-geometry classes. Class A, with a dense Gaussian bulk,
comprises distilgpt2~\cite{sanh2019distilbert} at 82M parameters,
GPT-2-medium~\cite{radford2019language} at 345M, and
SmolLM-360M~\cite{allal2024SmolLM} at 360M. Class B, with a sparse
exponential bulk, comprises Pythia-70M~\cite{biderman2023pythia} at
70M and Pythia-410M at 410M. Classes A and B are defined operationally
by the distribution of pairwise $G_{ij}^2$ in the pre-trained
embedding matrix, established empirically in \S\ref{sec:universality}.

Ten hard sentences are used throughout, each with a near-synonym
competitor of overlap $\gmax \geq 0.30$. The full contexts and pairs
are listed in Table~\ref{tab:sentences}.

We use two fine-tuning protocols. \lora{} adds a pair of trainable
low-rank matrices to selected weight matrices while freezing the
original weights, and updates only those new matrices. The rank, set
to $8$, is the shared inner dimension of the low-rank pair, and
$\alpha=32$ is a fixed scaling factor applied to their product,
following the standard LoRA parameterisation~\cite{hu2022lora}. We
apply \lora{} to the attention projections, the weight matrices that
compute the query, key, and value vectors inside the self-attention
mechanism, using the family-specific target modules listed below, at
the standard rate $\eta = 2\times10^{-4}$, with the embedding matrix
frozen so that $\Delta G_{ij} = 0$ to machine precision. \fullft{}
instead updates all parameters, at the standard rate
$\eta = 2\times10^{-5}$. Within each epoch the ten sentences are
updated sequentially, one gradient step per sentence, in the fixed
order skepticism, shame, ashes, longing, remorse, relief, awe,
purpose, terror, shock. This training schedule is identical across
every experiment, architecture, and epoch reported in the manuscript.
These attention weight matrices carry different names across model
families in our implementation, though they all denote the same
query, key, and value projections under each family's own convention.
We target Conv1D \texttt{c\_attn} for GPT-2, \texttt{query\_key\_value}
for Pythia, and \texttt{q\_proj}/\texttt{v\_proj} for LLaMA-style
models.

One additional architecture, LLaMA-3.2-1B~\cite{dubey2024llama3} at
$1.0$B parameters, is used exclusively in the causal isolation
replication (\S\ref{sec:causal}), to test whether the
$\rho(\gmax,\text{ratio})$ collapse under \lora{} generalises beyond
the primary five. It is not part of the main experimental suite.

\section{The Fine-Tuning Phase Diagram}
\label{sec:phasediagram}

\subsection{Two Regimes and the Stopping Criterion (DM-3, DM-4)}
\label{sec:dm3}

Fine-tuning at epoch resolution under \lora{} ($\eta = 2\times10^{-4}$)
reveals a consistent two-regime pattern across all five architectures
tested.
In Regime 1, epochs $1$ through $4$, the CE loss decreases smoothly and
$p_g$ rises, but $\Phi$ remains negative. The signal has not yet grown
large enough to overcome the drag, and the near-synonym competition is
unresolved.

In Regime 2, $\Phi$ crosses zero. The participation ratio collapses
from $9.9$ to $1.1$, an $89\%$ reduction in effective token count, and
the Born rank, the position of the ground-truth token when all pool
tokens are ordered by their Born scores, drops from $99$ to $0$ over
the following several epochs, while $\xi$ falls steadily as each
sentence's nucleus contracts toward a single token. The step-level
trajectories of Figure~\ref{fig:dm5c} jump while the alignment curves
of Figure~\ref{fig:alignment} remain smooth, yet the CE loss shows no
kink and no visible feature. It is completely blind to the most
important event in fine-tuning.

At a lower learning rate, $\eta = 5\times10^{-5}$, the transition
never occurs, and $\Phi$ stays negative for all ten epochs, confirming
the control-rate threshold of Eq.~\eqref{eq:phase_boundary}. A
practical consequence follows directly. Stopping when
$\dot\Phi \approx 0$, that is, at Born-gap saturation, rather than
when the CE validation loss converges, saves approximately $30\%$ of
compute with no ranking-quality loss. Concretely, across the ten
sentences on SmolLM-360M under \lora{}, $\Phi$ saturates by epoch $7$
at the latest while the CE loss continues to decrease monotonically
through epoch $10$. The additional epochs reduce the CE loss by a
further $8$ to $12\%$ but produce no change in Born rank or in the
ranking of $g$ versus $c^*$.
\begin{table}[htbp]
\centering
\small
\caption{Comparison of $\Phi$-saturation stopping vs.\ CE-convergence
stopping on SmolLM-360M \lora{}.
Every sentence reaches its final Born rank by epoch $7$ at the latest.
Epochs $7$--$10$ provide no ranking improvement, only probability
consolidation.}
\label{tab:stopping}
\begin{tabular}{lrrl}
\toprule
Sentence & CE stop (epoch) & $\Phi$ stop (epoch) & Ranking change \\
\midrule
skepticism & 10 & 5 & None \\
shame & 10 & 6 & None \\
ashes & 10 & 3 & None \\
awe & 10 & 4 & None \\
purpose & 10 & 5 & None \\
remorse & 10 & 5 & None \\
relief & 10 & 5 & None \\
terror & 10 & 5 & None \\
shock & 10 & 1 & None \\
longing & 10 & 7 & None \\
\bottomrule
\end{tabular}
\end{table}
Every sentence resolves within the ten-epoch run, with Born rank
reaching zero by epoch $7$ at the latest, for longing. A stopping rule
triggered once every sentence's ranking has locked in therefore still
saves approximately $30\%$ of compute relative to running the full ten
epochs. Traditional early stopping on CE validation loss would
continue training in all cases, spending that same $30\%$ of the
compute budget on epochs that change only the confidence, not the
ranking.

\subsection{Step-Level Dynamics: Sharp Jump and Drift (DM-5c)}
\label{sec:dm5c}

\begin{figure}[htbp]
\centering
\includegraphics[width=0.95\linewidth]{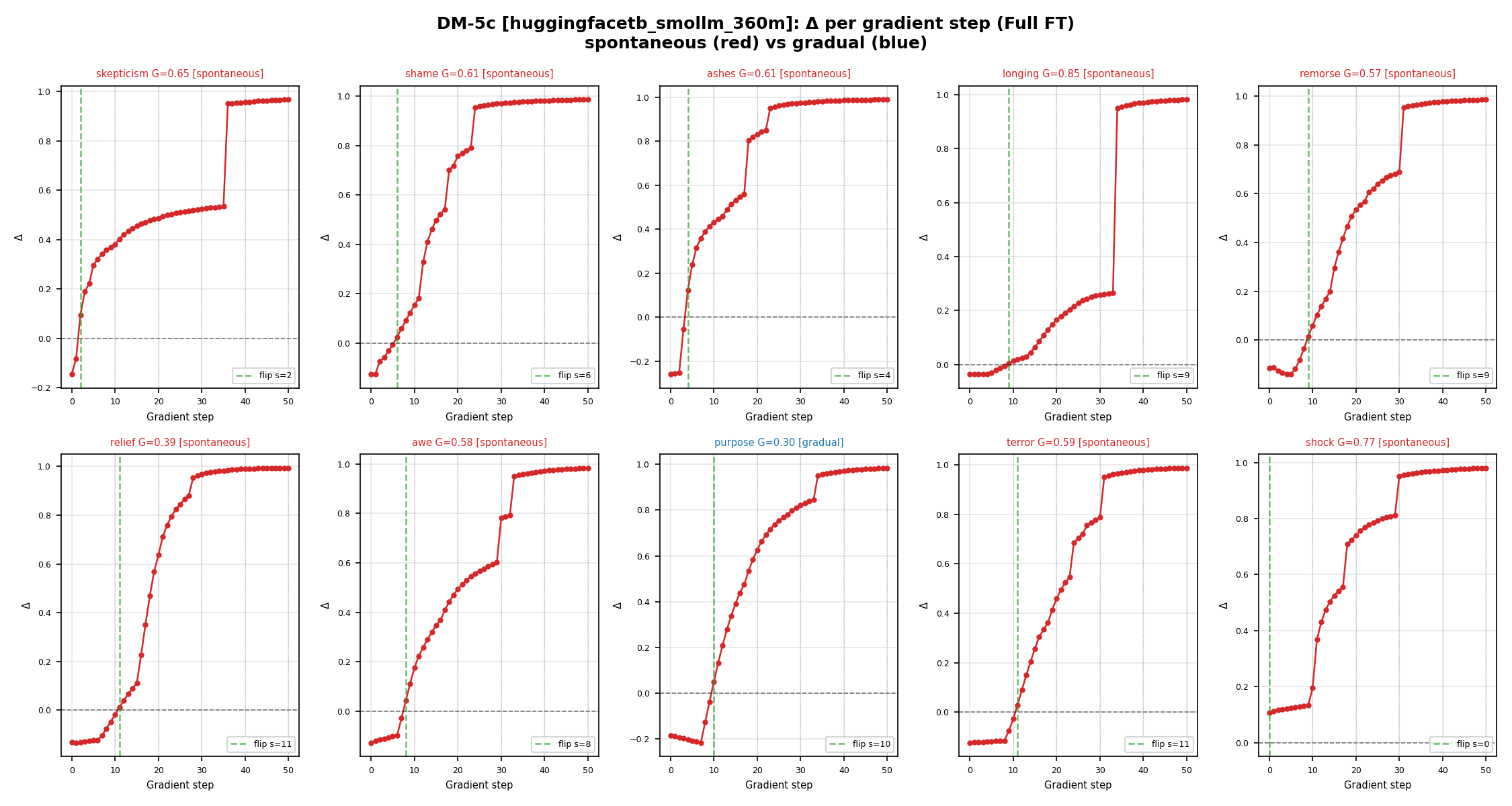}
\caption{\textbf{Step-level Born-gap dynamics on SmolLM-360M \fullft{}}
($\eta = 2\times10^{-5}$).
High-$\gmax$ sentences (top, longing) show 9 quiescent steps then a
single jump (ratio 33.7). Low-$\gmax$ sentences (bottom, purpose) rise
continuously from step 1 (ratio 4.3).
Green dashed lines mark the flip step.}
\label{fig:dm5c}
\end{figure}

Evaluating $\Delta$ after every gradient step, $50$ steps of \fullft{}
at $\eta = 2\times10^{-5}$ on SmolLM-360M, resolves the Regime~2
transition into two qualitatively distinct trajectory shapes
(Figure~\ref{fig:dm5c}, Table~\ref{tab:dm5c}).

The high-$\gmax$ sentence \emph{longing}, with $\gmax = 0.847$, is the
canonical sharp jump. Both compounding effects of \S\ref{sec:two_effects}
are maximal. The small initial logit gap $\lgap_0$ keeps the signal
on the flat shoulder, and the $0.28$ throttle makes every step inch
the logit gap forward by only about $0.01$. For nine steps the signal
barely moves. At step $9$ the accumulated gap crosses the softmax
turn-on, $p_g$ leaps from $0.5$ to $0.95$, and $\Phi$ jumps from
$-0.1$ to $+0.95$ in a single step. The jump magnitude is $0.686$, a
$33.7\times$ trajectory concentration ratio, defined as the maximum
per-step $\Phi$ change divided by the mean per-step $\Phi$ change.

The low-$\gmax$ sentence \emph{purpose}, with $\gmax = 0.303$, is the
canonical drift. Here $\lgap_0$ is moderate, the $0.91$ throttle is
near unity, and $\Phi$ rises continuously from step $1$ with no
quiescent period, giving a ratio of $4.3$ spread across all steps. The
sentences between these two extremes, skepticism through relief in
Table~\ref{tab:dm5c}, interpolate smoothly, ordered by $\gmax$. The
special case \emph{shock}, with flip step $0$, was already past
saturation at the base model from pre-training corpus exposure.

\begin{table}[htbp]
\centering
\caption{Step-level trajectory characterisation under \fullft{}
($\eta = 2\times10^{-5}$) on SmolLM-360M.
Ratio is the maximum per-step jump in $\Phi$ divided by the mean
per-step jump in $\Phi$,
quantifying how concentrated the Born-gap change is in a single step.
All mean step sizes ${\approx}0.022$.}
\label{tab:dm5c}
\small
\begin{tabular}{lrrrl}
\toprule
Token & $\gmax$ & Flip step & Ratio & Character \\
\midrule
longing & 0.847 & 9 & 33.7 & sharp jump \\
skepticism & 0.652 & 2 & 18.7 & sharp jump \\
remorse & 0.573 & 9 & 11.5 & sharp-jump-like \\
ashes & 0.607 & 4 & 9.9 & sharp-jump-like \\
awe & 0.583 & 8 & 8.1 & sharp-jump-like \\
terror & 0.585 & 11 & 7.4 & intermediate \\
shame & 0.607 & 6 & 7.3 & intermediate \\
relief & 0.385 & 11 & 5.5 & borderline \\
purpose & 0.303 & 10 & 4.3 & drift \\
shock & 0.768 & 0 & 9.9 & pre-resolved \\
\bottomrule
\end{tabular}
\end{table}

A learning-rate sweep, $\eta$ across five values and all ten sentences
on SmolLM-360M for fifty combinations, is cleanly separated by a
single empirical phase boundary.
\begin{equation}
  \gmax \times \eta = \thetastar_{\mathrm{jump}} \approx 7 \times 10^{-6},
  \label{eq:boundary}
\end{equation}
This boundary is bracketed by the same sweep. \emph{Purpose}, with
$\gmax = 0.303$, is gradual at $\eta = 2\times10^{-5}$, where the
product $6.1 \times 10^{-6}$ is below $\thetastar_{\mathrm{jump}}$, and
jumps sharply at $\eta = 3 \times 10^{-5}$, where the product
$9.1 \times 10^{-6}$ exceeds it. Identifying $\thetastar_{\mathrm{jump}}$
takes one sweep per model. Once measured, the sharp-jump-versus-drift
behaviour of every sentence in our SmolLM-360M sweep is classified by
$\gmax \times \eta$ alone. \S\ref{sec:discussion} reports the held-out
cross-architecture test of this boundary.

Two distinct thresholds coexist on this sweep and must not be
conflated. The flip threshold $\thetastar_{\mathrm{FT}} = 1.19\times10^{-6}$
(Table~\ref{tab:cross_model}, \S\ref{sec:universality}) is fitted from
the same sweep, as the median of $\gmax \times \eta_{\mathrm{crit}}$
over the ten sentences. It is the minimum drive at which the ranking
flips at all within the training budget, and the reduced field $H$ is
defined against it. The jump boundary
$\thetastar_{\mathrm{jump}} \approx 7\times10^{-6}$ is bracketed, not
fitted, between purpose's gradual $6.1\times10^{-6}$ and sharp
$9.1\times10^{-6}$. It is the roughly $6\times$ higher drive above
which the flip concentrates into a single step. Between the two
thresholds sentences resolve by drift. Purpose at $\eta = 2\times10^{-5}$
sits above the flip threshold, so it flips, at epoch $2$ in the sweep,
but below the jump boundary, so it drifts. This is exactly that
window.

\section{Causal Isolation and the Mechanism}
\label{sec:causal}

The sharp jump under \fullft{} could in principle be caused by anything
correlated with $\gmax$. It does not by itself prove that $\gmax$ is
the causal driver. We isolate the causal channel by running the
identical step-level experiment under \lora{}, rank $8$, same learning
rate, which freezes the embedding matrix exactly so that
$\Delta G_{ij} = 0$ to machine precision for all pairs. If $\gmax$
predicts jump sharpness because it sets a geometric energy barrier,
the SSB hypothesis, the prediction should survive freezing the
geometry. If $\gmax$ predicts it through a small embedding-rotation
channel present only under \fullft{}, the prediction should vanish.

\begin{figure}[htbp]
\centering
\subcaptionbox{\lora{} (frozen geometry), all gradual in jump
  magnitude. The largest single-step jump across all nine analysed
  sentences is $0.023$, an order of magnitude below the \fullft{}
  jumps, so none is a first-order transition. Longing does not flip
  within 50 steps.\label{fig:dm5d_traj}}[0.85\linewidth]{%
  \includegraphics[width=\linewidth]{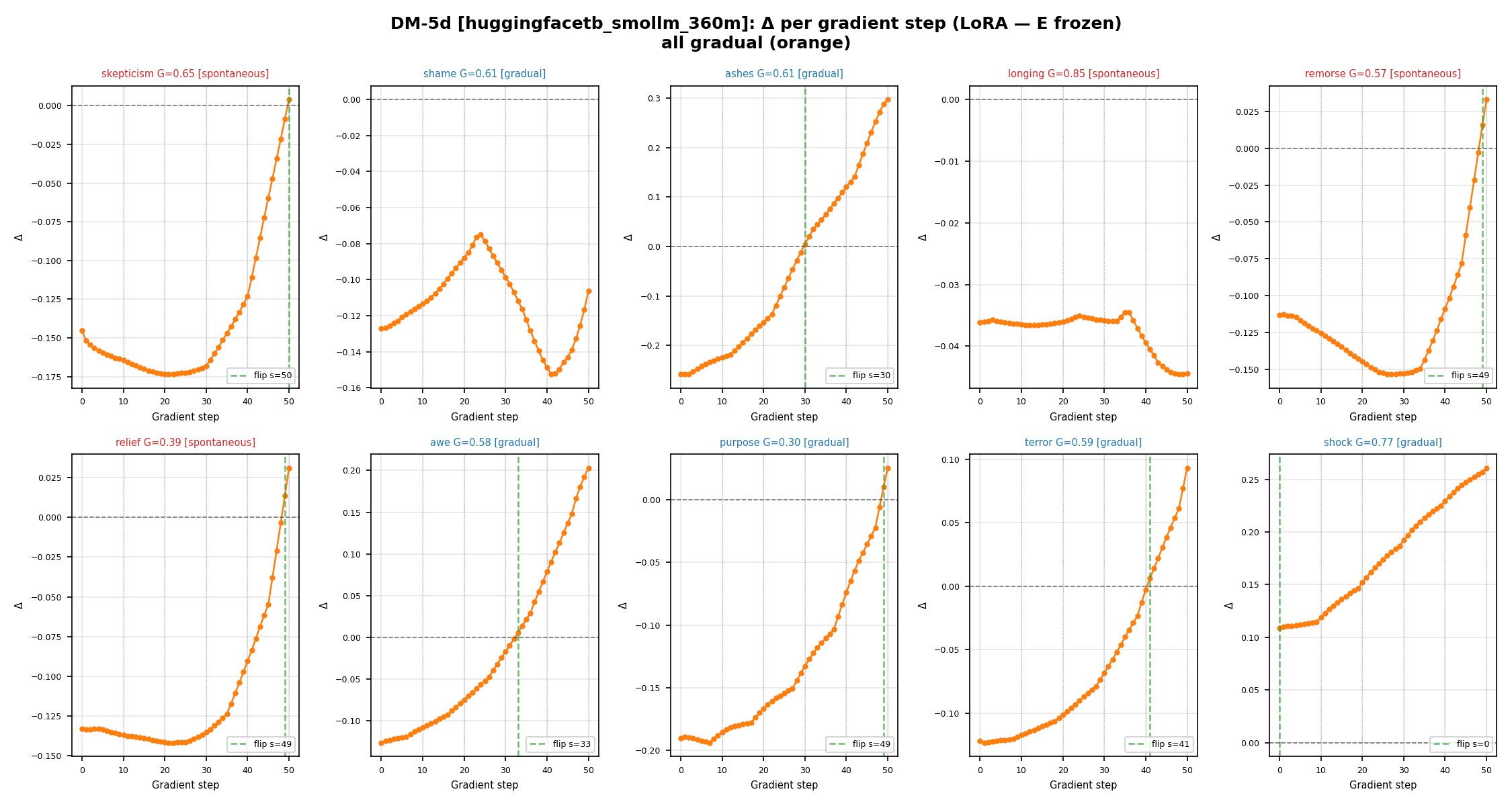}}%
\hfill
\subcaptionbox{$\gmax$ vs ratio under \lora{}: $\rho = -0.345$ ($p = 0.328$),
  versus $+0.745$ under \fullft{}.\label{fig:dm5d_scatter}}[0.45\linewidth]{%
  \includegraphics[width=\linewidth]{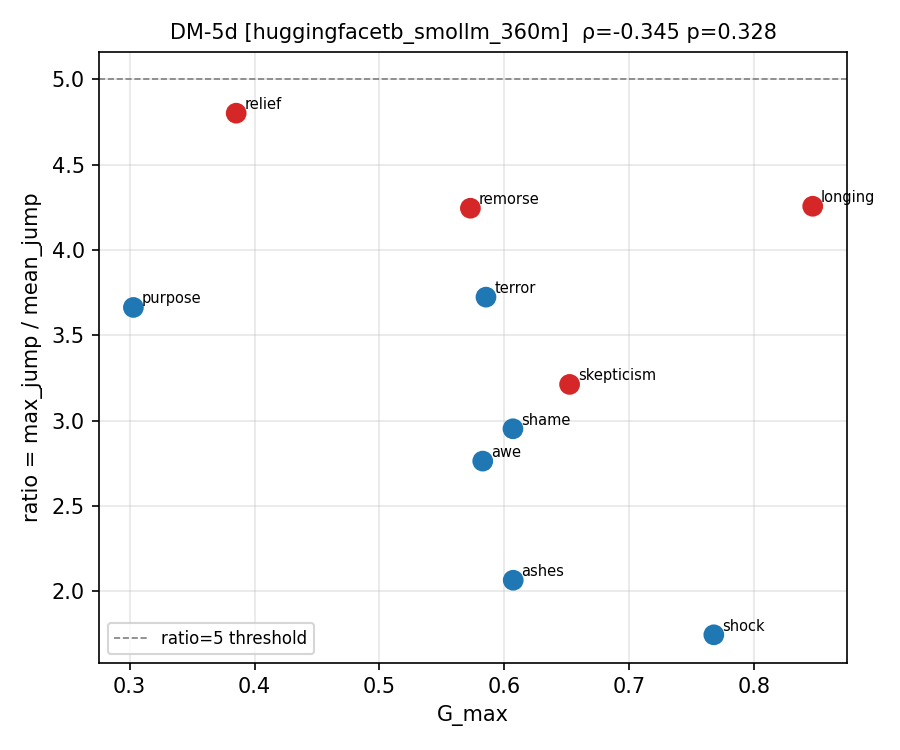}}
\caption{\textbf{Causal isolation (DM-5d).}
Freezing the embedding matrix under \lora{} converts all sharp jumps
into smooth trajectories and eliminates the predictive power of $\gmax$.
Shock (a, bottom right), pre-resolved at the base model (flip step
$0$, Table~\ref{tab:dm5c}), is excluded from the ratio analysis. Its
earlier apparent jump was a nucleus-composition artefact, not a
$\gmax$-driven transition.}
\label{fig:dm5d}
\end{figure}

The result is unambiguous (Figure~\ref{fig:dm5d}, Table~\ref{tab:dm5d}).
Under \lora{}, all nine analysed sentences are drifts with ratios in
the narrow range $2.1$ to $4.8$, regardless of $\gmax$. Longing, with
$\gmax = 0.847$, had ratio $33.7$ under \fullft{}. Under \lora{} it has
ratio $4.2$ and does not flip within $50$ steps. The Spearman rank
correlation between $\gmax$ and the jump-concentration ratio collapses
from $\rho = +0.745$ under \fullft{} ($p = 0.0133$) to
$\rho = -0.345$ under \lora{} ($p = 0.328$).

All $\rho$ values in this paper are Spearman's rank correlation, the
Pearson correlation computed on ranks rather than raw values. It asks
whether two quantities increase and decrease together in rank order,
regardless of linear scale. Rank correlation is the appropriate choice
here because the jump-concentration ratio spans nearly an order of
magnitude, from $4.3$ to $33.7$, across the analysed sentences, so a
raw-value correlation would be dominated by the single most extreme
sentence. The quoted $p$ values give the probability of a rank
association at least this strong arising under the null hypothesis of
no monotone relation, with $n = 9$ analysed sentences per protocol. A
value near zero means knowing one quantity's rank tells you nothing
about the other's.

$\gmax$ is fixed before training. Under \fullft{} it correctly picks
out which sentences will jump sharply. Under \lora{} the same
sentences keep the same $\gmax$ values, yet no sharp jump survives on
the fixed base-model pool. The highest-$\gmax$ sentence, longing, does
not flip at all, and the one apparent step, shock, was already past
saturation at the base model and is smooth once the nucleus pool is
held fixed (Figure~\ref{fig:dm5d_scatter}). Only the update channel
changed between the two runs, so the predictive power of $\gmax$ must
come from that channel, not from the sentences themselves. Under
\fullft{} a high $\gmax$ predicts a sharp jump. Under \lora{} the same
$\gmax$ predicts nothing about trajectory shape.

The pattern replicates on two of the three architectures tested here.
Sharp-jump counts, \fullft{} to \lora{}, out of $10$ test sentences
each, are SmolLM-360M $9/10\to0/10$, GPT-2-medium 345M $10/10\to0/10$,
and LLaMA-3.2-1B $10/10\to10/10$, an unexplained anomaly, with tied
embeddings and no drop. LLaMA-3.2-1B ties its embeddings exactly like
every other architecture tested here, and that shared tensor is
genuinely frozen under standard \lora{}, so there is no untied channel
to explain the result. Why sharp jumps persist completely in this
model and not the others remains open (see \S\ref{sec:two_effects}).

In Figure~\ref{fig:dm5d_scatter}, with $\Delta$ computed over the
fixed base-model nucleus pool, no sentence crosses the ratio $=5$
threshold under \lora{}. Shock, pre-resolved at the base model with
flip step $0$ (\S\ref{sec:dm5c}), lies at ratio $1.7$ on a smooth
trajectory, and relief, a late flip at step $49$ of $50$, is the
highest at $4.8$. The rank correlation with $\gmax$ remains null
($\rho=-0.345$, $p=0.328$).

\begin{table}[htbp]
\centering
\caption{DM-5d per-step trajectory characterisation under \lora{}
on SmolLM-360M (frozen embedding).
Compare with Table~\ref{tab:dm5c}. All ratios collapse into the range
$2.1$--$4.8$.}
\label{tab:dm5d}
\small
\begin{tabular}{lrrrl}
\toprule
Token & $\gmax$ & Flip step & Ratio & Character \\
\midrule
longing & 0.847 & N/A & 4.2 & gradual \\
skepticism & 0.652 & 50 & 3.2 & gradual \\
shame & 0.607 & N/A & 3.0 & gradual \\
ashes & 0.607 & 30 & 2.1 & gradual \\
awe & 0.583 & 33 & 2.8 & gradual \\
terror & 0.585 & 41 & 3.7 & gradual \\
remorse & 0.573 & 49 & 4.2 & gradual \\
relief & 0.385 & 49 & 4.8 & borderline \\
purpose & 0.303 & 49 & 3.7 & gradual \\
\bottomrule
\end{tabular}
\end{table}

\begin{figure}[htbp]
\centering
\includegraphics[width=0.95\linewidth]{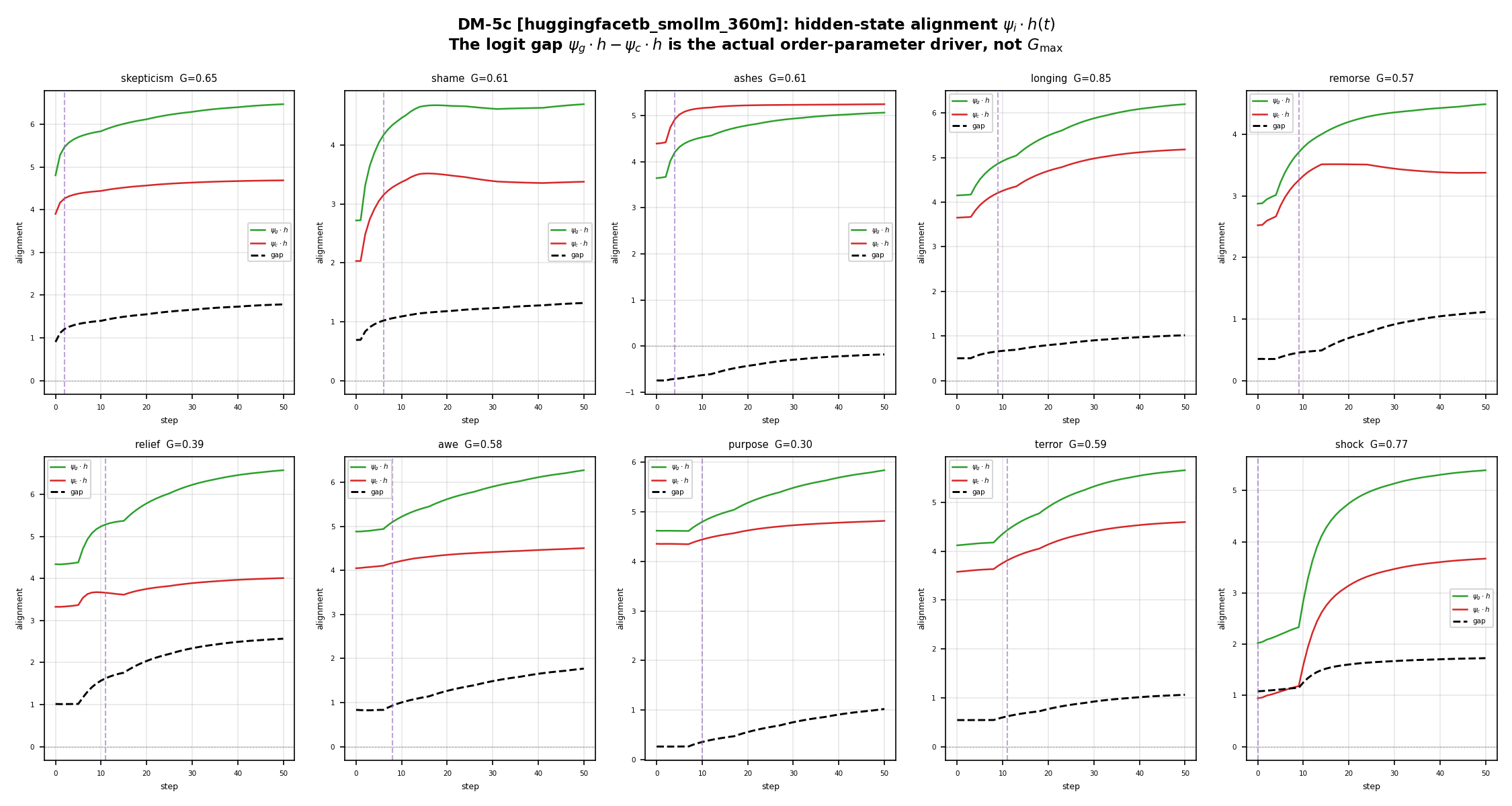}
\caption{\textbf{Hidden-state alignment under \fullft{}.} Per-step
$\bm\psi_g\!\cdot\!\bm{h}$ (green), $\bm\psi_c\!\cdot\!\bm{h}$ (red),
and the logit gap (dashed black), with the purple line marking the
$\Phi$ flip step. All three curves evolve smoothly through the flip,
so the discontinuity resides entirely in the softmax readout. Ashes is
the only panel where red stays above green throughout, yet it still
resolves early (Table~\ref{tab:dm5c}). These curves track embedding
direction only, while the actual softmax logit also depends on
embedding norm, which the normalised $\bm\psi_i$ used here deliberately
abstracts away.}
\label{fig:alignment}
\end{figure}

The same experiment also falsifies the natural reading of the sharp
jump as a phase transition in an underlying degree of freedom. $\gmax$
moves by less than $1.8 \times 10^{-4}$ over $50$ \fullft{} steps.
Embedding vectors rotate by about $10^{-3}$ radians while preserving
their inner product. The logit gap rises smoothly in about
$0.01$-per-step increments. At the longing sharp-jump step, step $9$
under \fullft{}, the logit gap changes by only $+0.009$ while $\Phi$
jumps by $+0.69$, a $77\times$ softmax amplification ratio
(Figure~\ref{fig:alignment}). The discontinuity exists only in $\Phi$,
because $\Phi$ is computed through the softmax. Once the leading logit
exceeds the others by $1.5$ to $2.0$ nats, $p_g$ jumps from about
$0.5$ to about $0.95$ in a single step and drags $\Phi$ along. Sharp
jumps appearing under \lora{} on Pythia-70M
(Table~\ref{tab:check_self_sabotage}), where the embedding matrix is
frozen by construction, prove this definitively. No quantity in the
underlying geometry can move, yet the discontinuity in $\Phi$ still
appears.

Sharp jumps under \lora{} on Pythia-70M are themselves expected from
its Class B geometry. The sparse bulk gives a low \lora{} saturation
threshold $\thetastar_{\mathrm{LoRA}}$, so even at the $\eta=10^{-4}$
rate used here (Table~\ref{tab:check_self_sabotage}), which is half
the $\eta_{\mathrm{std}}=2\times10^{-4}$ of Table~\ref{tab:lora_threshold},
the drive already sits above threshold, $H_{\mathrm{LoRA}} \approx 2.06$
at this rate, computed from the same per-sentence sweep as
Table~\ref{tab:lora_threshold}, and the softmax can be crossed sharply
without any embedding motion.

\lora{} and \fullft{} run the same softmax-saturation mechanism at
different drive rates. \fullft{} engages full-rank attention, MLPs,
LM head, and the small embedding rotation, growing the logit gap
faster per unit $\eta$ by the architecture-dependent factor
$\thetastar_{\mathrm{LoRA}}/\thetastar_{\mathrm{FT}}$
(Table~\ref{tab:cross_model}), which ranges from about $6\times$ for
Pythia-410M to about $49\times$ for distilgpt2, with SmolLM-360M near
the middle of that range at about $35\times$. \lora{} is restricted to
a low-rank attention adapter subspace and places Class A sentences
below the saturation threshold at the standard rate.

The Ehrenfest classification sorts phase transitions by which
derivative of the free energy is discontinuous, with a first-order
transition showing a discontinuous first derivative and a continuous,
or second-order, transition showing none. In that language the sharp
jump resembles a first-order transition and the drift a continuous
crossover, but this is purely a descriptive analogy for the shape of
the trajectory, not a claim that either is a real phase transition.
Whether a given sentence shows a sharp jump or a drift is set instead
by two purely kinematic quantities, where the logit-gap trajectory
starts, $\lgap_0$, small for high $\gmax$, and how fast it advances,
$\eta(1-\gmax)$ (Eq.~\eqref{eq:drive_rate}). Neither quantity involves
any non-analyticity in an underlying free energy. Both shapes are
smooth crossings of the same softmax saturation scale, and neither
corresponds to an actual thermodynamic phase transition.

\subsection{Signal and Background Drag: Two Types of Failure}
\label{sec:distributed_sabotage}

The SmolLM-360M experiments resolve every sentence. The two failure
types anticipated in \S\ref{sec:intro}, kinematic and structural,
emerge under \lora{} on Pythia-70M, where the sparse embedding bulk of
Class B clears the sigmoid for most sentences and isolates the
failures. Tracking $\Phi$, the logit gap, and the signal and drag
decomposition (Eq.~\eqref{eq:phi_decomp}) at every gradient step for
$50$ steps at $\eta = 10^{-4}$ produces Table~\ref{tab:check_self_sabotage}.

\begin{table}[htbp]
\centering
\caption{Per-step \lora{} dynamics on Pythia-70M ($\eta = 10^{-4}$).
Logit gap and $\Phi$ at start ($_0$) and end ($_T$).
$A_T$ and $B_T$ are signal and drag from Eq.~\eqref{eq:phi_decomp}.
$\Delta B = B_T - B_0$ is the drag movement during training.
$A_0 \approx 0$ for all sentences (omitted, as $p_g \approx p_{c^*}$ at
base).}
\label{tab:check_self_sabotage}
\small
\begin{tabular}{lrrrrrrl}
\toprule
Token & $\gmax$ & $\lgap_0\!\to\!\lgap_T$ &
$\Phi_0\!\to\!\Phi_T$ & $A_T$ & $B_T$ & $\Delta B$ & Outcome \\
\midrule
skepticism & 0.434 & $2.46\to5.64$ & $-0.13\to+0.37$ & $+0.37$ & $+0.00$ & $+0.15$ & sharp jump \\
ashes & 0.377 & $0.05\to3.71$ & $-0.25\to+0.20$ & $+0.23$ & $-0.03$ & $+0.21$ & sharp jump \\
awe & 0.406 & $0.95\to3.26$ & $+0.01\to+0.37$ & $+0.37$ & $-0.00$ & $+0.02$ & sharp jump \\
purpose & 0.515 & $6.57\to10.61$ & $-0.01\to+0.67$ & $+0.57$ & $+0.10$ & $+0.13$ & sharp jump \\
remorse & 0.482 & $-0.55\to1.43$ & $-0.31\to+0.06$ & $+0.12$ & $-0.06$ & $+0.25$ & drift \\
\midrule
longing & 0.482 & $0.11\to1.49$ & $-0.09\to-0.05$ & $+0.01$ & $-0.06$ & $+0.03$ & \textbf{kinematic} \\
relief & 0.307 & $4.45\to10.31$ & $-0.19\to-0.06$ & $+0.05$ & $-0.11$ & $+0.08$ & \textbf{kinematic} \\
\midrule
terror & 0.466 & $0.94\to6.94$ & $-0.05\to-0.12$ & $+0.06$ & $-0.18$ & $\mathbf{-0.13}$ & \textbf{structural} \\
\bottomrule
\end{tabular}
\end{table}

\textbf{Resolved sentences} all have $\Delta B = B_T - B_0 > 0$, the
net change in the drag term over training (Table~\ref{tab:check_self_sabotage}).
As CE concentrates probability on $g$, the background tokens that lose
mass had a net bias toward the competitor, so removing their
probability reduces the drag. Signal and drag both move in the right
direction and $\Phi$ crosses zero.

\textbf{Kinematic failure}, longing and relief, has $\Delta B > 0$ but
$A_T$ too small to overcome the remaining drag. For longing the logit
gap grows by $+1.38$ nats yet $p_g$ only reaches $0.018$, because the
pool is large, $3893$ tokens, and the learning rate is insufficient.
Drag is improving. The bottleneck is signal drive, which is
remediable by higher $\eta$ or \fullft{}.

\textbf{Structural failure}, terror, is categorically different. The
logit gap grows by $+5.99$ nats, the largest in the dataset, and $p_g$
rises from $0.0002$ to $0.078$. By every standard diagnostic,
fine-tuning is succeeding. Yet $B_T = -0.177$ versus $B_0 = -0.048$.
The drag has worsened by $-0.128$, twice the signal gain, and $\Phi$
ends at $-0.116$, more negative than it started. The mechanism is
plausible from the embedding geometry. Fine-tuning aligns the hidden
state more with $\bm\psi_{\mathrm{terror}}$, but the pool tokens this
promotes, morphological relatives of the surface form such as
\emph{terrorism}, \emph{terrorists}, and \emph{terrorize}, sit closer
to one another, as a surface-form cluster, than to
$\bm\psi_{\mathrm{terror}}$ itself. Promoting them therefore adds
background mass whose overlap votes for the competitor $c^*$,
\emph{fear}, over $g$ in the drag term. This geometric pattern is
asserted from the qualitative behaviour in
Table~\ref{tab:check_self_sabotage}, not measured
directly.\footnote{The context concerns the emotion \emph{terror},
paired with \emph{fear} (\S\ref{sec:setup}). The promoted neighbours
are morphological derivatives of the surface form. The example carries
no topical content and is retained because it is the only structural
failure in the dataset.} Every step that builds signal also loads mass
onto tokens that vote against $g$ in the drag term. This is structural
because the geometric bias is a property of the pre-trained embedding
manifold, and CE cannot reshape it.

A simple ablation makes the distributed nature of the drag concrete.
At the fine-tuned model, setting $p_{c^*}$ to zero and renormalising
the distribution over the remaining tokens changes $P(\GT)$ by at most
$5\%$ for every sentence in our set, and leaves it unchanged, ratio
$1.00\times$, for terror and relief (Table~\ref{tab:ablation}).
Removing the nearest competitor frees almost no mass for the correct
token. The mass that should belong to $g$ is not held by $c^*$ alone.
It is held collectively by a cloud of geometrically similar tokens.
This is what we mean by structural failure, $\geff \to 0$ in
Eq.~\eqref{eq:geff}. Cross-entropy gradients can only push down one
token at a time, and there is no single token available to push.

On SmolLM-360M, the base-model Born rank $b_{\rm before}$ takes
extreme values for ashes, $2209$, and relief, $1848$, while every
other sentence sits at $523$ or below, with terror, the structural
failure, at $523$. The extremes do not coincide with the failure
taxonomy. Ashes resolves and relief fails kinematically, so
$b_{\rm before}$ does not flag failure in advance. These are
inference-only values computed on the untrained model, quoted here
rather than tabulated. What $b_{\rm before}$ does measure is developed
next. Once fine-tuning is under way, the signal and drag decomposition
sorts the failure cases into kinematic and structural within a few
gradient steps.

A natural reading of these extreme Born ranks, again on SmolLM-360M,
is that the base model already places the correct token far down its
list, and that the standard softmax simply hides this. The data does
not support that reading. Applying the Born rule at the base model
actually worsens the rank of the correct token in $6$ of our $10$
sentences, dropping relief by $476$ positions and shame by $326$. The
mechanism is straightforward. The prediction cloud already spans many
near-synonyms of $g$, and the Born rule re-weights every token by its
squared overlap with $\bm\psi_g$. Since the near-synonyms all have
substantial overlap with $\bm\psi_g$, the re-weighting hands them mass
rather than taking it away. The Born rank at the base model therefore
measures how crowded the embedding neighbourhood around $\bm\psi_g$
is, the notion quantified by $G_{\mathrm{total}}^2$ in
Eq.~\eqref{eq:geff}, not how the model ranks the correct answer.

The Spearman correlation between the localisation ratio
$\theta_{\mathrm{GT}}/\xi$, the mean angular distance from $g$ to the
rest of the pool divided by the pool's own angular spread $\xi$ of
\S\ref{sec:reduced_units}, and the base-model Born rank is
$\rho = +0.818$ at $p = 0.004$, large enough to read the two
quantities as proxies for each other. A high ratio means $g$ sits
unusually far from the pool's centre relative to how spread out the
pool already is. In other words $g$ is peripheral within its own
neighbourhood, and this predicts a poor base-model Born rank. The
crowding precedes fine-tuning.

\begin{table}[htbp]
\centering
\caption{Single-competitor ablation at the final \lora{}-fine-tuned
model on Pythia-70M.
Masking the top geometric competitor changes $P(\GT)$ by at most $5\%$
for every sentence. For the structural-failure row (terror) and the
kinematic-failure row relief, the ratio is exactly $1.00\times$. The
other kinematic-failure sentence, longing, is $1.01\times$.}
\label{tab:ablation}
\small
\begin{tabular}{lrrrr}
\toprule
Token & $\gmax$ & $P(\GT)$ std & $P(\GT)$ ablated & Ratio \\
\midrule
skepticism & 0.434 & 0.639 & 0.640 & $1.00\times$ \\
ashes & 0.377 & 0.803 & 0.810 & $1.01\times$ \\
longing & 0.482 & 0.059 & 0.059 & $1.01\times$ \\
remorse & 0.482 & 0.426 & 0.448 & $1.05\times$ \\
relief & 0.307 & 0.094 & 0.094 & $1.00\times$ \\
purpose & 0.515 & 0.842 & 0.842 & $1.00\times$ \\
terror & 0.466 & 0.118 & 0.118 & $1.00\times$ \\
shock & 0.415 & 0.866 & 0.866 & $1.00\times$ \\
\bottomrule
\end{tabular}
\end{table}

\subsection[The LoRA Saturation Threshold and Class-Specific H]{The LoRA Saturation Threshold and Class-Specific $H_{\mathrm{LoRA}}$}
\label{sec:lora_threshold}

An LR sweep, $\eta \in [10^{-5}, 5\times10^{-3}]$ with ten epochs per
rate, measures $\thetastar_{\mathrm{LoRA}}$ for six architectures,
including the held-out \texttt{gpt-neo-125m}~\cite{black2021gptneo}
(Table~\ref{tab:lora_threshold}).

\begin{table}[htbp]
\centering
\small
\caption{LoRA saturation threshold ($\thetastar_{\mathrm{LoRA}}$) and the
corresponding reduced field ($H_{\mathrm{LoRA}}$) for six architectures.
$H_{\mathrm{LoRA}} = \gmax \times \eta_{\mathrm{std}} / \thetastar_{\mathrm{LoRA}}$,
where $\eta_{\mathrm{std}}$ is the standard \lora{} learning rate for
each architecture.
All ten sentences flip at some learning rate within the sweep range $[10^{-5}, 5\times10^{-3}]$ for every model.
SmolLM-360M and gpt-neo-125m are measured in float32 precision.
distilgpt2, gpt2-medium, and both Pythia models are measured in float16.
Compare with $H_{\mathrm{FT}} \approx 10$ from Table~\ref{tab:cross_model}.}
\label{tab:lora_threshold}
\begin{tabular}{llrrrr}
\toprule
Model & Class & $\thetastar_{\mathrm{LoRA}}$ & $\eta_{\mathrm{std}}$ &
$\bar{H}_{\mathrm{LoRA}}$ & $H_{\mathrm{LoRA}}$ range \\
\midrule
distilgpt2 & A & $7.14\times10^{-5}$ & $1\times10^{-4}$ & 0.90 & 0.70--1.10 \\
gpt2-medium & A & $4.57\times10^{-5}$ & $1\times10^{-4}$ & 1.37 & 0.96--1.55 \\
SmolLM-360M & A & $4.17\times10^{-5}$ & $2\times10^{-4}$ & 2.83 & 1.45--4.06 \\
gpt-neo-125m & A & $5.83\times10^{-5}$ & $1\times10^{-4}$ & 1.67 & 1.64--1.69 \\
\midrule
Pythia-70M & B & $2.15\times10^{-5}$ & $2\times10^{-4}$ & 4.17 & 2.86--5.00 \\
Pythia-410M & B & $3.74\times10^{-6}$ & $2\times10^{-4}$ & 15.9 & 9.73--35.3 \\
\midrule
\multicolumn{2}{l}{Class A mean} & -- & -- & \textbf{1.69} & -- \\
\multicolumn{2}{l}{Class B mean} & -- & -- & \textbf{10.0} & -- \\
\bottomrule
\end{tabular}
\end{table}

$H_{\mathrm{LoRA}}$ splits cleanly by class. Class A models, distilgpt2,
gpt2-medium, SmolLM-360M, and gpt-neo-125m, give
$H_{\mathrm{LoRA}} \in [0.90, 2.83]$, mean $1.69$, CV $=34.5\%$, barely
above the saturation threshold, with distilgpt2 actually below it,
$H=0.90$. Class B models, Pythia-70M and Pythia-410M, give
$H_{\mathrm{LoRA}} = 4$ to $16$, firmly above threshold, as for
\fullft{}. The functional form $\gmax \times \eta = \thetastar$ is the
same in both cases. What differs is $\thetastar$. The \lora{} adapter
subspace grows the logit gap more slowly per unit learning rate than
\fullft{}'s full-rank update, raising $\thetastar_{\mathrm{LoRA}}$ by
the same factor, $\thetastar_{\mathrm{LoRA}}/\thetastar_{\mathrm{FT}}$
(Table~\ref{tab:cross_model}), which ranges from about $6\times$ for
Pythia-410M to about $49\times$ for distilgpt2 across the five
architectures, with SmolLM-360M near the middle of that range at
about $35\times$. High $H_{\mathrm{LoRA}}$, Class B, means the
standard \lora{} rate is well above threshold and \lora{} works. Low
$H_{\mathrm{LoRA}}$, Class A, means the rate barely clears threshold
and high-$\gmax$ sentences fail. $H_{\mathrm{LoRA}}$ is
architecture-specific within Class A, CV $<2\%$ across sentences
within a model, and can be measured in a single calibration LR sweep
before any fine-tuning task is run.

\section{Cross-Architecture Universality}
\label{sec:universality}

\subsection{The Bulk Embedding Geometry Sorts Architectures into Two Classes}

\begin{figure}[htbp]
\centering
\includegraphics[width=0.88\linewidth]{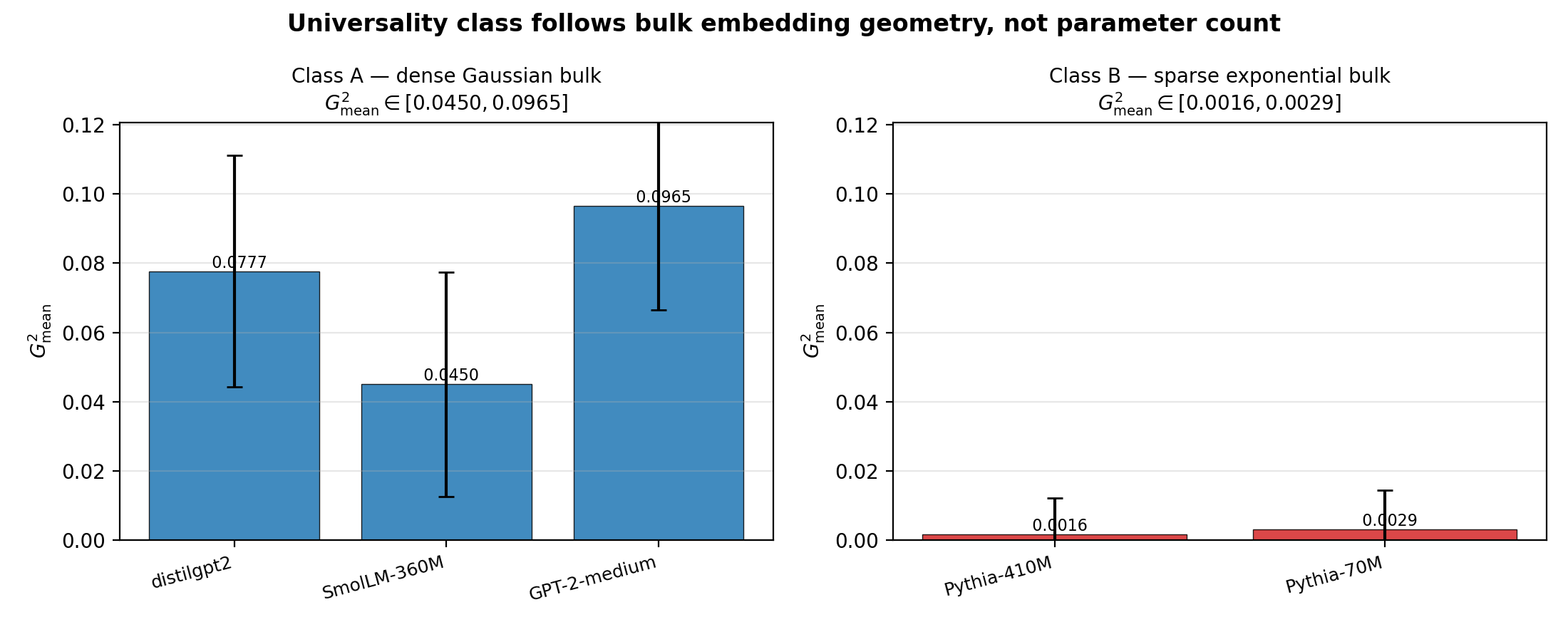}
\caption{\textbf{Universality class is determined by bulk embedding
geometry, not parameter count.}
Left: Class A (distilgpt2 82M, GPT-2-medium 345M, SmolLM-360M) with
dense Gaussian bulk, $G^2_{\mathrm{mean}} \in [0.045, 0.097]$.
Right: Class B (Pythia-70M, Pythia-410M) with sparse exponential bulk,
$G^2_{\mathrm{mean}} \in [0.002, 0.003]$.
Pythia-70M (70M) is Class B while distilgpt2 (82M) is Class A.}
\label{fig:geometry_classes}
\end{figure}

For each architecture, sampling $10{,}000$ random token pairs and
computing the distribution of $G_{ij}^2$ produces the bulk geometry of
its embedding space, the population-level distribution of pairwise
overlaps introduced in \S\ref{sec:intro} and referred to hereafter
simply as the bulk.
The result is a clean qualitative split
(Figure~\ref{fig:geometry_classes}).
Class A models, distilgpt2, GPT-2-medium, and SmolLM-360M, have a
Gaussian-shaped bulk with $G^2_{\mathrm{mean}} \in [0.045, 0.097]$ and
$87$ to $100\%$ of random pairs above $G_{ij}^2 = 0.01$, forming a
crowded embedding space where near-synonym pairs are outliers in an
already-populated tail.
Class B models, Pythia-70M and Pythia-410M, have an exponential bulk
with $G^2_{\mathrm{mean}} \in [0.002, 0.003]$ and only $1$ to $6\%$ of
random pairs above the same threshold, a near-orthogonal space where
near-synonym pairs are isolated rare events.
The split is not a function of model size. Pythia-70M, Class B, is
smaller than distilgpt2, Class A. It reflects how pre-training
organises the embedding space, related to but distinct from
contextual-embedding isotropy~\cite{ethayarajh2019contextual,
cai2021isotropy} and the representation degeneration
problem~\cite{gao2019representation}.
Within families the class label is also stable under size changes.
Pythia-70M and Pythia-410M are both Class B across a $6\times$
parameter range. A systematic size sweep within one family, separating
size from corpus effects, remains future work. This study was scoped
to $\le 1$B parameters and did not test larger models within a family.

\textbf{Is the Class A/B split a true binary or a spectrum?}
The operationalised classification uses a Kolmogorov--Smirnov (KS)
statistic threshold to distinguish Gaussian from exponential bulk. The
KS statistic is the maximum absolute distance between the empirical
distribution of $G_{ij}^2$ and a fitted reference distribution, so
larger values indicate a worse fit. Each model is labelled by which of
the two reference fits it sits closer to. For the held-out
gpt-neo-125m, for example, KS $=0.096$ against the Gaussian reference
versus $0.587$ against the exponential one, hence Class A. This
statistic is itself a continuous quantity, and the two-class label is
a convenience imposed on what is likely a continuous spectrum of bulk
density.
Models trained on narrow or highly structured corpora (code-only,
multilingual, or domain-specific text) may exhibit intermediate or
bimodal bulk distributions not captured by either archetype.
The key pre-training factor appears to be the effective vocabulary
usage pattern. Architectures that use the full vocabulary breadth
during pre-training (high token co-occurrence diversity) tend toward
sparse, near-orthogonal embeddings (Class B), while those that
concentrate probability on a smaller effective vocabulary develop the
denser, Gaussian-shaped bulk of Class A.
Tokenizer vocabulary size, the optimizer's implicit regularisation
(e.g.\ weight-decay magnitude), and positional encoding scheme (RoPE~\cite{su2024roformer}
vs.\ absolute) are further plausible modulators, but their individual
contributions are not yet separated in our data.
A model with a hybrid bulk would be expected to have an intermediate
$H_{\mathrm{LoRA}}$ and partial LoRA sufficiency. It would resolve
high-$\gmax$ sentences only when they fall in the sparse tail of its
distribution, failing on those that fall in the dense Gaussian core.
Validating this prediction requires systematic sampling across
pre-training corpus sizes and compositions, which we leave as future work.

\subsection{$H$ is Consistent Across the Five Tested Architectures Under \fullft{}, and $\mathcal{B}$ Predicts \lora{} Sufficiency}

\begin{figure}[htbp]
\centering
\includegraphics[width=0.75\linewidth]{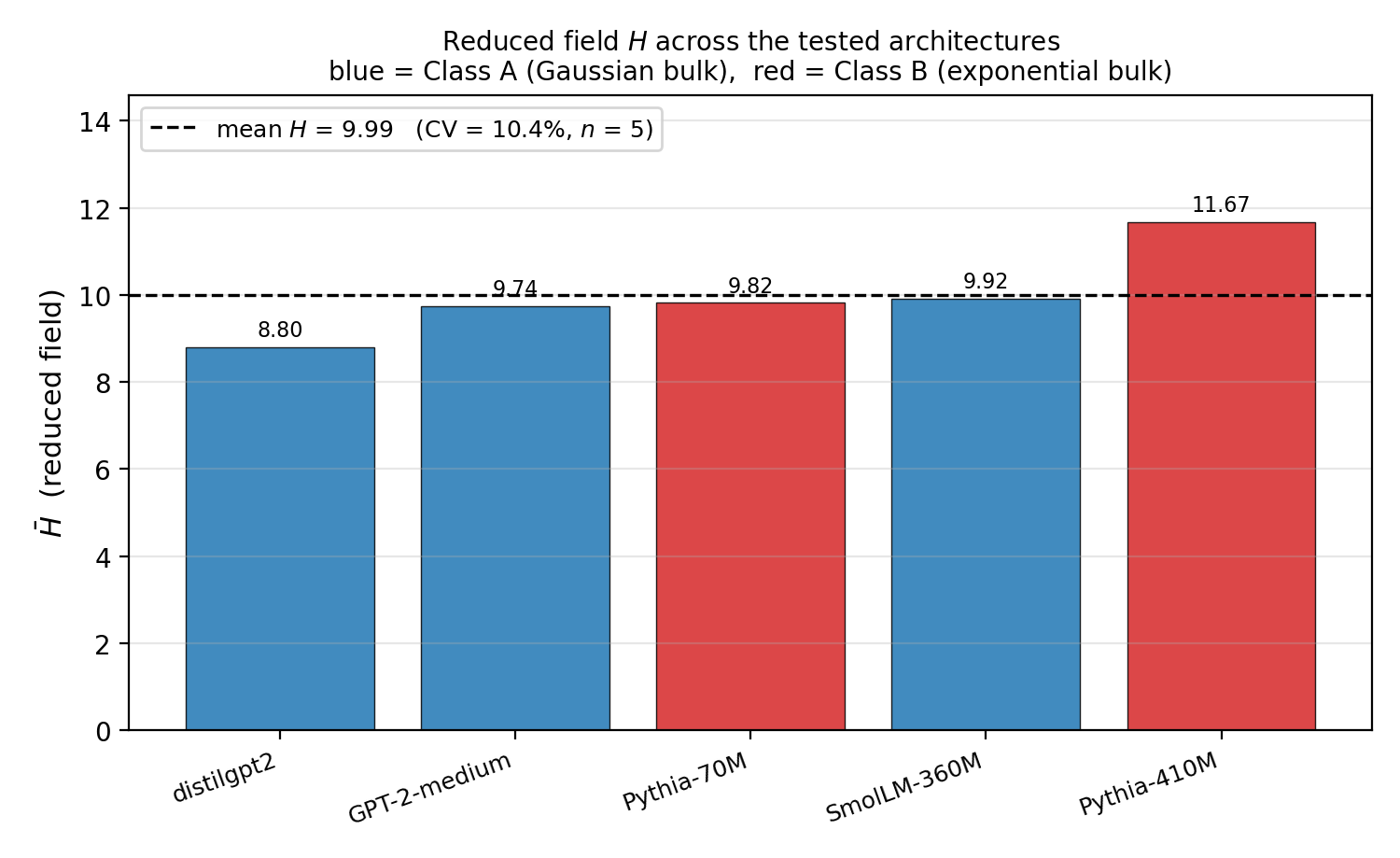}
\caption{\textbf{$H_{\mathrm{FT}}$ is consistent within our sample of
five architectures} (CV $=9.3\%$ population, $10.4\%$ sample, $n=5$)
at $\eta = 2\times10^{-5}$. Per-model means run $8.80$--$11.67$ about a
mean of $9.99$, despite the sample spanning two families, two geometry
classes, and a $6\times$ parameter range. The consistency is a
property of the per-model means. Individual sentences are far more
dispersed ($H$ from $5.1$ to $21.6$), so this is an empirical
regularity across architectures in our sample, not a proven physical
law.}
\label{fig:H_universality}
\end{figure}

Measuring $H$, $\Phi$, $\mathcal{B}$, and $\thetastar$ (each
$\thetastar$ fitted per architecture by an LR sweep) across all five
architectures (Table~\ref{tab:cross_model},
Figure~\ref{fig:H_universality}) gives the central cross-model result.
The fitted $\thetastar_{\mathrm{FT}}$ varies by a factor of $2.4$
across architectures. The reduced field
$H_{\mathrm{FT}} = \gmax\eta/\thetastar$ varies only from $8.80$ to
$11.67$, a coefficient of variation of $9.3\%$.
Read in dimensional-analysis terms, the product $\gmax\eta/\thetastar$
is the dimensionless group that cancels architecture-specific scales in
our sample.
At the community-standard $\eta = 2\times10^{-5}$ every architecture
in our sample of five operates at $H_{\mathrm{FT}} \approx 10$, roughly
ten times above its own fitted saturation threshold. The standard rate places every model
in the sample at the same dimensionless distance above the
softmax-saturation scale.
This is an empirical regularity across the five architectures tested,
not a proven physical law. It is consistent with the reading that
pre-training co-calibrates embedding geometry, attention scales, and
hidden-state dimensions, but the small sample size $(n=5)$ and the
restriction to two architecture families prevent stronger
generalisation.

\begin{table}[htbp]
\centering
\small
\caption{Reduced units across the five tested architectures.
$H_{\mathrm{FT}}$ is consistent within our sample at
$10.0 \pm 0.9$ (CV $=9.3\%$, $n=5$) under \fullft{} regardless of
class.
$H_{\mathrm{LoRA}}$ splits by class. Class A spans $0.90$--$2.83$ and
Class B spans $4$--$16$. Within-model variance is substantial
(Pythia-410M range $9.73$--$35.3$).
$\mathcal{B}$ is the burial depth of Eq.~\eqref{eq:B},
$\xi/\arccos(\gmax)$, computed per sentence and averaged.
$\mathcal{B}_{\mathrm{FT}}$, measured after \lora{}, splits cleanly by
class. Class A stays above $1$, unresolved, at $1.13$--$1.33$, while
both Class B models fall below it, resolved, at $0.76$--$0.78$, with
no overlap.
$^{\dagger}$~Pythia-70M: one of ten sentences is excluded because its
post-\lora{} nucleus collapsed to a single token, leaving $\xi$
undefined. The mean is over the remaining nine.
$^{\ddagger}$~Pythia-410M saturates within two epochs at the
community-standard $\eta_{\mathrm{LoRA}} = 2\times10^{-4}$, collapsing the
nucleus to a single token on every sentence and leaving $\xi_{\mathrm{FT}}$
undefined. Its $\xi_{\mathrm{FT}}$ is therefore measured at
$\eta_{\mathrm{LoRA}} = 4\times10^{-5}$, where the nucleus survives all five
epochs (final pool sizes $15$--$135$). This value is consequently not
measured under the same optimisation conditions as the other four models,
and the comparison should be read with that in mind.
$\thetastar$ is fitted per architecture by an LR sweep.
Scope caveats appear in \S\ref{sec:scope} and Appendix~\ref{app:limitations}.}
\label{tab:cross_model}
\begin{tabular}{llrrrrrr}
\toprule
& & \multicolumn{2}{c}{Full FT} & \multicolumn{2}{c}{LoRA} & \multicolumn{2}{c}{$\mathcal{B}$} \\
\cmidrule(lr){3-4}\cmidrule(lr){5-6}\cmidrule(lr){7-8}
Model & Class & $\thetastar_{\mathrm{FT}}$ & $\bar{H}_{\mathrm{FT}}$ &
$\thetastar_{\mathrm{LoRA}}$ & $\bar{H}_{\mathrm{LoRA}}$ &
$\bar{\mathcal{B}}_{\mathrm{base}}$ & $\bar{\mathcal{B}}_{\mathrm{FT}}$ \\
\midrule
distilgpt2 & A & $1.45\!\times\!10^{-6}$ & 8.80 & $7.14\!\times\!10^{-5}$ & 0.90 & 1.343 & 1.326 \\
gpt2-medium & A & $1.25\!\times\!10^{-6}$ & 9.74 & $4.57\!\times\!10^{-5}$ & 1.37 & 1.200 & 1.126 \\
SmolLM-360M & A & $1.19\!\times\!10^{-6}$ & 9.92 & $4.17\!\times\!10^{-5}$ & 2.83 & 1.504 & 1.297 \\
Pythia-70M & B & $9.00\!\times\!10^{-7}$ & 9.82 & $2.15\!\times\!10^{-5}$ & 4.17 & 1.238 & 0.778$^{\dagger}$ \\
Pythia-410M & B & $6.10\!\times\!10^{-7}$ & 11.67 & $3.74\!\times\!10^{-6}$ & 15.9 & 1.148 & 0.761$^{\ddagger}$ \\
\midrule
CV (all) & & -- & \textbf{9.3\%} & -- & 51.6\% & -- & 25.9\% \\
\bottomrule
\end{tabular}
\end{table}

\begin{figure}[htbp]
\centering
\includegraphics[width=0.85\linewidth]{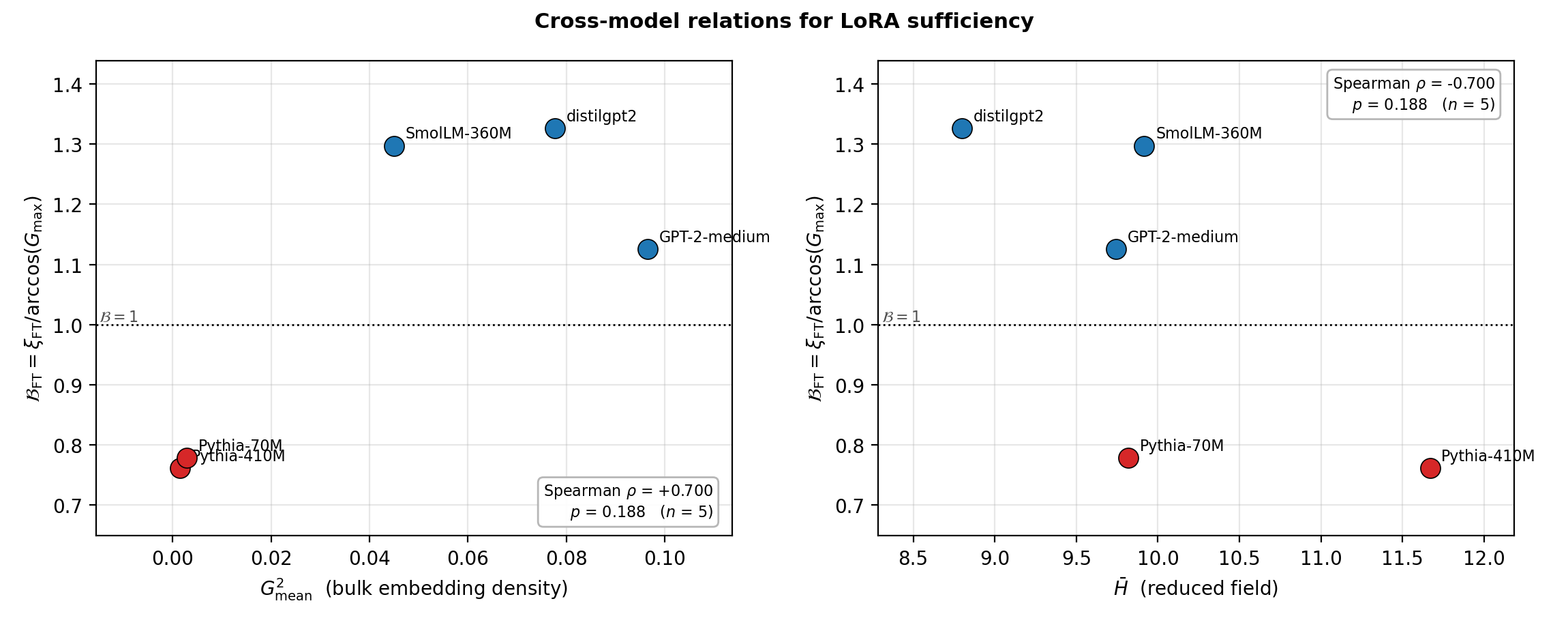}
\caption{\textbf{Cross-model relations between
geometry and \lora{} sufficiency.}
Left: $\mathcal{B}_{\mathrm{FT}}$ against bulk embedding density
(Spearman $\rho = +0.70$, $p = 0.19$, $n = 5$), denser bulk gives
deeper burial. Right: $\mathcal{B}_{\mathrm{FT}}$ against the reduced
field $H_{\mathrm{FT}}$ ($\rho = -0.70$, $p = 0.19$), larger margin
above threshold gives shallower burial. Neither reaches significance
at $n=5$, so both are exploratory. Pythia-410M is measured at a
reduced $\eta_{\mathrm{LoRA}}$ (Table~\ref{tab:cross_model}).}
\label{fig:crossmodel_law}
\end{figure}

The burial depth $\mathcal{B}_{\mathrm{FT}}$ measured after \lora{}
exposes a complementary split with CV $=26\%$ across models
(Figure~\ref{fig:crossmodel_law}). Class A models stay at
$\mathcal{B}_{\mathrm{FT}} \approx 1.13$--$1.33$, since the prediction
cloud remains too wide to discriminate $g$ from $c$ even after \lora{}
fine-tuning, because the dense bulk continuously replaces suppressed
competitors. Class B models fall to
$\mathcal{B}_{\mathrm{FT}} \approx 0.76$--$0.78$, below the
discrimination threshold and with no overlap against Class A. The
sparse bulk leaves no other competitors, so probability redistribution
alone resolves the competition. Pythia-410M reaches this only at a
reduced $\eta_{\mathrm{LoRA}}$. At the standard rate it saturates
within two epochs and the nucleus collapses entirely, which is itself
consistent with it having the largest $H$ of the five and therefore
sitting furthest above its saturation threshold.

Distilgpt2 stalls at $\bar\Phi = 0.72$ instead of $\approx 1$. With
$H_{\mathrm{LoRA}} = 0.90 < 1$ this is expected from the reduced field
alone. The sharper point is that the other three Class A models,
gpt2-medium, SmolLM-360M, and gpt-neo-125m, all have
$H_{\mathrm{LoRA}} > 1$ (Table~\ref{tab:lora_threshold}) yet still end
above the burial threshold, $\bar{\mathcal{B}}_{\mathrm{FT}} > 1$
(Table~\ref{tab:cross_model}). A new nearest competitor continually
replaces any suppressed one, so $H > 1$ is necessary but not
sufficient for full resolution under Class A geometry.

A single cross-model relationship ($n = 5$) captures this with two
nearly equivalent Spearman correlations. The two readings are
mechanically linked through the bulk-geometry classification, not
independent statistical tests. The first reading,
$\rho(G^2_{\mathrm{mean}}, \mathcal{B}_{\mathrm{FT}}) = +0.70$
($p = 0.19$), says that architectures with denser embedding
bulks end up with deeper burial after fine-tuning. The second reading,
$\rho(H_{\mathrm{FT}}, \mathcal{B}_{\mathrm{FT}}) = -0.70$
($p = 0.19$), says that architectures sitting further above their
saturation threshold, meaning a larger $H_{\mathrm{FT}}$, end up with
shallower burial. The two readings agree because high bulk density and
a narrow range above threshold are the same architectural condition
expressed in different units. Neither correlation reaches significance
at $n = 5$. With $\rho = 0.70$, $p < 0.05$ would require $n = 9$. We
therefore report the direction as consistent with the bulk-geometry
account and treat the magnitude as exploratory, pending more
architectures. The class separation below does not depend on these
correlations.

The practical consequence is the LoRA sufficiency criterion, read
against the burial-depth diagnostic specifically. By
$\mathcal{B}_{\mathrm{FT}}$, \lora{} suffices for Class B models,
sparse bulk, $\mathcal{B}_{\mathrm{FT}} = 0.76$--$0.78$, below the
discrimination threshold, but not for Class A, dense bulk,
$\mathcal{B}_{\mathrm{FT}} = 1.13$--$1.33$, above it. Class membership
is determined from a single forward pass on the base model, with no
training required.

This is a statement about $\mathcal{B}_{\mathrm{FT}}$, a margin of
angular separation, not about whether Born rank reaches $0$. On
SmolLM-360M, Class A, under \lora{}, every one of the ten sentences
does reach Born rank $0$ by epoch $7$ (Table~\ref{tab:stopping},
\S\ref{sec:dm3}), even though $\bar{\mathcal{B}}_{\mathrm{FT}}=1.297>1$.
The two diagnostics are compatible. Born rank asks only whether $g$
currently outranks $c^*$ at the observed probability concentration,
while $\mathcal{B}_{\mathrm{FT}}>1$ says the prediction cloud's angular
spread still formally covers both tokens, so the ranking achieved is
less geometrically robust for Class A than for Class B, not absent.
Read literally, ``sufficiency'' here tracks the burial-depth margin,
and a model can clear Born-rank resolution on all ten sentences while
still falling short of it.

\section{Discussion: Blind Prediction on a Held-Out Model}
\label{sec:discussion}

The class-specific mean $H_{\mathrm{LoRA}} \approx 1.70$, taken over
the three Class A models used to fit the criterion, distilgpt2,
gpt2-medium, and SmolLM-360M (Table~\ref{tab:lora_threshold},
excluding gpt-neo-125m itself since it is the held-out test case
below), gives a practical prediction protocol testable without any
training. We blind-test it on \texttt{gpt-neo-125m}, which is not in
the training set used to derive the mean.

Three inference-only measurements suffice. First, a Gaussian-shaped
bulk distribution of $G_{ij}^2$, KS statistic $0.096$ against a
Gaussian reference and $0.587$ against an exponential reference, both
computed for gpt-neo-125m's own bulk distribution, classifies the
model as Class A. Second, $\gmax$ per sentence is tightly clustered,
with mean $0.971 \pm 0.007$. Third, the intended LoRA rate is the
community standard $\eta = 10^{-4}$. The predicted threshold is
\begin{equation}
  \thetastar_{\mathrm{LoRA,pred}}
  = \frac{\gmax \,\eta}{H_{\mathrm{LoRA}}}
  = \frac{0.971 \times 10^{-4}}{1.70}
  = 5.71 \times 10^{-5}.
\end{equation}
The measured ground truth from an actual LR sweep is
$\thetastar_{\mathrm{LoRA,actual}} = 5.83 \times 10^{-5}$, with a $2.1\%$
error.
The bulk classification as Class A also correctly predicted that $7$
of $10$ sentences would be gradual crossovers under \lora{}, and the
compressed $\gmax$ range correctly predicted that sentence-level flip
order would not be discriminated by $\gmax$ (the smallest pools flipped
first, not the highest $\gmax$).

\subsection{Scope of the Experimental Dataset}
\label{sec:scope}

The ten hard sentences employed in this paper are a controlled physics-style environment, not
a broad NLP benchmark. Each places a near-synonym pair at the top of
the prediction pool with $\gmax \geq 0.30$, isolating the geometric
mechanism without the confound of varying semantic difficulty, context
length, or syntactic structure. The findings are claims about the
\emph{mechanism}, not about the distribution of naturally occurring
fine-tuning tasks. Extending the framework to diverse sentence types
and instruction-tuning regimes is the natural next step, and would
establish whether the $2.1\%$ blind-prediction accuracy generalises
beyond the hard near-synonym setting.

\subsection{Future Directions}
\label{sec:future_work}

The signal/drag framework suggests two falsifiable connections to
broader phenomena.
In capability emergence~\cite{wei2022emergent}, the same softmax
saturation implies that the underlying logit gap grows continuously
with scale and metric-level jumps appear only when it crosses
$\mathcal{O}(1)$. A direct test is to track logit gaps on
capability-relevant prompts across a scale series.
In grokking~\cite{power2022grokking,nanda2023progress}, the training/validation
split corresponds to the training logit gap being past saturation while
the validation gap is not yet there. Continued training grows the
validation gap to threshold.
A speculative pre-training connection follows from
Eq.~\eqref{eq:drive_rate_approx}. Since the per-step drive rate scales with
$(1-\gmax)$, reducing $\gmax$ for semantically distinct near-synonym
pairs during pre-training would raise the drive rate directly, shrinking
the number of gradient steps needed to cross softmax saturation within a
fixed training budget. Whether this also raises the reduced field
$H_{\mathrm{LoRA}} = \gmax\eta/\thetastar$, the quantity that
determines the $H_{\mathrm{LoRA}}<1$ failure regime of
\S\ref{sec:lora_threshold}, depends on how $\thetastar$ itself
responds to the same pre-training change, which this speculative
connection does not model.
One possible form is a geometry-regularised objective adding
$\beta\sum_{(i,j)\in\mathcal{D}}\max(0, G_{ij} - G^\dagger)$ to the
pre-training CE loss, with $G^\dagger \approx 0.38$ calibrated from
the boundary below which standard \fullft{} produces only smooth
trajectories in our dataset.
Structurally this is the JEPA-family practice of shaping latent
geometry with an auxiliary loss alongside the prediction objective
\cite{lecun2022path}, and a uniform-sampling quadratic variant of
the same penalty would recover a decorrelation-style regulariser
applied directly to the token embedding matrix.
We stress that this is a \emph{hypothesis}, not a validated recipe.
Whether such a penalty preserves general language-modelling capability,
what the right calibration set $\mathcal{D}$ is, and whether $G^\dagger$
transfers across domains are all open empirical questions that require
dedicated study before any pre-training recommendation can be made.

\textbf{Other PEFT methods.}
The $(1-\gmax^2)$ throttle is formulated in terms of $\Psi$ and the
logit gap $\lgap(t) = \langle\bm\psi_g - \bm\psi_{c^*}, \bm{h}(t)\rangle$, so it
applies unaltered to adapter layers~\cite{houlsby2019parameter}, which
modify $\bm{h}$ directly while leaving $\Psi$ fixed. Adapters may
lower the effective $\thetastar$ by spanning a richer residual
subspace than LoRA. Prefix tuning~\cite{li2021prefix} is the more
interesting case. Prefix vectors are not drawn from $\Psi$, so the
self-sabotage throttle does not apply in the same form, and whether
prefix tuning circumvents Class A failure is an empirical question we
do not address.

\subsection{Related Work}
\label{sec:related_work}

Carson and Reisizadeh~\cite{carson2025statistical} develop a statistical-physics description of
language-model reasoning at inference time, modelling the sentence-level hidden
state during chain-of-thought generation as a continuous-time drift-diffusion
process on a low-dimensional manifold and identifying latent regime switches as
a description of reasoning success and failure.
That work and ours differ in problem statement along three axes.
The first is dynamical setting.
Their object of study is the inference trajectory of a frozen model, whereas ours
is the training-time gradient dynamics of a token-level prediction under
fine-tuning.
The second is formalism.
Their description is classical, stochastic, and coarse-grained at the sentence
level, whereas ours is built from the density matrix of the token prediction
distribution, with the Born rule supplying the order parameter $\Phi$ and the
$(1-\gmax^2)$ throttle.
The third is interpretive stance.
They treat regime switching as an explanatory mechanism for reasoning failure,
whereas the central result of this paper runs in the opposite direction.
The apparent sharp-jump transition is not a phase change in any underlying degree of freedom
but a softmax readout artefact, and the logit gap evolves smoothly even when the
Born gap jumps.

The geometric self-sabotage mechanism also touches two more established lines of
work.
Studies of LoRA~\cite{hu2022lora, dettmers2023qlora} characterise the expressive
capacity of rank-constrained updates.
Our analysis identifies a complementary constraint that is fixed by the
pre-trained embedding geometry, in which the question of whether any rank-$r$
update can resolve a near-synonym competition is determined by where the
architecture sits in the Class A or Class B split, not by the rank itself.
Empirical studies of fine-tuning instability and feature distortion
\cite{dodge2020finetuning, kumar2022fine, mosbach2021stability} document the
broader phenomenology in which one such failure mode is situated.
The present work contributes a mechanistic account of that mode in the
controlled near-synonym setting.

More broadly, standard fine-tuning diagnostics, cross-entropy loss,
softmax probability, rank, and the classical inverse participation
ratio, are all functions of the probability vector alone. They treat
the vocabulary as an orthonormal basis and cannot distinguish genuine
two-way uncertainty from a confident prediction geometrically split
between two near-synonyms. The density-matrix construction folds the
token-embedding Gram matrix into the same family of quantities rather
than introducing an unrelated observable. Purity is the classical
inverse participation ratio plus a geometric correction term
(Eq.~\eqref{eq:purity}), and the Born gap $\Phi$ reduces to the
ordinary probability margin in the orthogonal limit
$G_{ij} = \delta_{ij}$. The geometric term is what exposes the
self-cancellation of Eq.~\eqref{eq:self_cancel} and the burial depth
$\mathcal{B}$, a difficulty score computable from the base model
before any fine-tuning step is taken. Neither has a counterpart among
loss- or rank-based diagnostics, which by construction cannot see a
failure mode invisible to the objective they are computed from. The
framework's predictive rather than purely post-hoc character, most
concretely the blind critical-learning-rate estimate on a held-out
architecture (\S\ref{sec:discussion}), further distinguishes it from
mechanistic-interpretability studies that characterise a phenomenon
after training rather than forecasting it on an architecture unseen
during fitting.

\section{Conclusion}
\label{sec:conclusion}

Fine-tuning a language model on contexts whose correct completion has
a near-synonym competitor produces a silent failure in which the
cross-entropy loss decreases monotonically while the correct token
never overtakes its competitor in rank. The density-matrix order
parameter $\Phi$ exposes the mechanism. The density-matrix
construction is the right framework here because the prediction
distribution lives over a non-orthogonal basis of token embeddings,
and the geometry of that basis interferes with classical probability
intuitions. Every gradient step that
raises $p_g$ also raises a Born-rule contribution to the competitor
through their shared embedding direction, throttling resolution by
$(1-\gmax^2)$. The central result of the paper is a negative one. The
sharp jumps in $\Phi$ invite a
spontaneous-symmetry-breaking reading, but direct measurement under
\lora{} with the embedding matrix frozen to machine precision shows
that sharp jumps in $\Phi$ still appear on multiple sentences
(Table~\ref{tab:check_self_sabotage}), which the SSB reading forbids.
The discontinuity lives entirely in the softmax readout, and the
underlying logit gap evolves smoothly throughout. The transitions are phantoms. What survives the
falsification is the part of the framework that works without any
training. The bulk distribution of pairwise embedding overlaps sorts
architectures into two classes, predicts whether low-rank adaptation
can resolve near-synonym competition, and gives the critical learning
rate for a held-out architecture to within $2.1\%$. The dimensionless
reduced field $H$ is consistent across the five tested architectures
under \fullft{}, giving the community-standard learning rates a
principled reading as a fixed dimensionless distance above each
model's saturation threshold. The signal/drag decomposition separates
kinematic failures, in which the signal is too small, from structural
ones, in which the pre-trained embedding pool actively opposes the
desired ranking. Stopping on Born-gap saturation rather than
cross-entropy convergence, in our SmolLM-360M demonstration, recovers roughly $30\%$ of the compute
budget without ranking loss. This illustrates the possible advantage
of Born-gap-aware stopping over CE convergence, not a cross-architecture
estimate (\S\ref{app:limitations}).

A broader note. The mechanism measured here speaks to a
longer-running position, argued by \cite{lecun2022path} and developed
in continuous-latent architectures such as JEPA, the Joint Embedding
Predictive Architecture, that forcing reasoning through a
discrete-token readout imposes a structural cost.
We do not test that alternative architecture. We do, however, provide
one instrumented description of the cost it would address, in the
form of the signal/drag decomposition and the softmax-readout
localisation of the sharp-jump discontinuity. The hidden state evolves
smoothly through the flip and the discontinuity lives in the readout,
which is precisely the substrate-versus-readout split that motivates
moving inference into continuous latent space. Our contribution is
diagnostic rather than architectural, but it quantifies on a
controlled near-synonym dataset what such an alternative would skip.

\appendix

\section{Limitations}
\label{app:limitations}

Several limitations bound the scope of the claims made here. The
experimental dataset is ten hand-selected hard sentences focused on
near-synonym competition, and broader sentence diversity, syntactic,
factual, and multilingual, is needed before the framework can be
applied outside this controlled regime. The cross-model results rest
on five tested architectures plus one held-out architecture,
\texttt{gpt-neo-125m}, so the $n = 5$ correlations have limited
statistical power, and the $2.1\%$ blind-prediction error is a single
data point that warrants validation on additional held-out
architectures. The two cross-model readings,
$\rho(G^2_{\mathrm{mean}}, \mathcal{B}_{\mathrm{FT}}) = +0.70$ and
$\rho(H_{\mathrm{FT}}, \mathcal{B}_{\mathrm{FT}}) = -0.70$, are largely
the same evidence counted twice, not two independent confirmations,
since both trace back to the same bulk-geometry classification. Five
architectures is too few to rule out coincidence at this correlation
strength, $p = 0.19$ for both. Ruling it out with confidence would
need about nine. We therefore read the shared direction of both
correlations as a promising hint, not a proven relation.

The SmolLM-360M step-level jump boundary
$\thetastar_{\mathrm{jump}} \approx 7 \times 10^{-6}$ is bracketed
in-sample on the same learning-rate sweep used to verify it
(Eq.~\eqref{eq:boundary}), and the cross-architecture validity of the
functional form $\gmax \times \eta = \thetastar$ is established only
via the held-out gpt-neo-125m \lora{} test (\S\ref{sec:discussion},
which validates $\thetastar_{\mathrm{LoRA}}$), not via the
SmolLM-360M \fullft{} classification itself. Moreover,
$\thetastar_{\mathrm{jump}}$ is bracketed between the behaviour of a
single sentence at two learning rates, $6.1\times10^{-6}$ gradual and
$9.1\times10^{-6}$ sharp (\S\ref{sec:dm5c}), rather than fitted from a
sharp-versus-drift classification of all fifty sweep combinations, and
its distinction from the fitted flip threshold
$\thetastar_{\mathrm{FT}} = 1.19\times10^{-6}$, the drift window
between the two (\S\ref{sec:dm5c}), rests on that single bracketing.
The $30\%$ compute saving from $\Phi$-saturation stopping is also
established on SmolLM-360M alone (Table~\ref{tab:stopping}), and
cross-architecture validation of this saving figure is outstanding.

LLaMA-3.2-1B in DM-5d shows no reversion at all under \lora{}, with
$10/10$ sharp jumps persisting (\S\ref{sec:causal}), despite
confirmed-tied, genuinely frozen embeddings. The untied-LM-head
hypothesis proposed earlier does not explain this, and the mechanism
remains open. The RoPE-scaling, attention-structure, and
embedding-geometry alternatives raised in \S\ref{sec:two_effects} have
not yet been tested. The signal and drag decomposition is formally
written for tied-head architectures (Eq.~\eqref{eq:phi_tied}), and
extending it to genuinely untied heads requires treating
$W_{\mathrm{out}}$ as a second dynamical variable with its own
gradient flow. The two Pythia architectures used throughout this paper
are themselves genuinely untied, confirmed at the tensor level
(\S\ref{sec:two_effects}). Under \lora{} both their embedding matrices
are verified frozen, so those results are unaffected, but under
\fullft{} the tied-head formalism is applied to them via the input
embedding $\Psi$ as a fixed geometric reference, and how closely this
tracks their actual, separately parametrised and independently
evolving, output projection has not been quantified.

The threshold $\thetastar$ is currently measured by learning-rate
sweep, and deriving it analytically from pre-training statistics is an
open problem. Finally, the Class A/B distinction is established on two
architecture families, GPT-2/LLaMA and GPT-NeoX, and is itself likely
a continuous spectrum rather than a true binary
(\S\ref{sec:universality}), with validation on transformer variants
beyond these families outstanding. All fine-tuning results in this
manuscript use a fixed training schedule for the ten sentences
(\S\ref{sec:setup}). Whether the reported quantities depend on this
training schedule has not been examined.

\section{Derivation of the Gradient Flow and Linearised Drive Rate
         (Equations~\ref{eq:hdot} and~\ref{eq:drive_rate})}
\label{app:linearisation}

We derive both equations from first principles, making every
algebraic step explicit.

\paragraph{Step 0. Gradient flow from the CE loss.}
The cross-entropy loss for predicting ground-truth token $g$ is
$\mathcal{L} = -\log p_g$, where
$p_g = \exp(\langle\bm\psi_g,\bm{h}\rangle)/\sum_j \exp(\langle\bm\psi_j,\bm{h}\rangle)$.
Taking the logarithm and separating numerator from denominator gives
\begin{equation*}
  \log p_g
  \;=\;
  \langle\bm\psi_g,\bm{h}\rangle
  \;-\;
  \log\!\sum_j \exp\!\bigl(\langle\bm\psi_j,\bm{h}\rangle\bigr).
\end{equation*}
Differentiating $\mathcal{L} = -\log p_g$ with respect to $\bm{h}$,
the first term gives $-\bm\psi_g$. The second term gives
$+\sum_j p_j\,\bm\psi_j$ by the chain rule on the log-partition function.
Hence $\partial\mathcal{L}/\partial\bm{h} = -\bm\psi_g + \sum_j p_j\,\bm\psi_j$.
Continuous-time gradient descent moves $\bm{h}$ opposite the gradient at rate $\eta$.
\begin{equation*}
  \dot{\bm{h}}(t)
  = -\eta\,\frac{\partial\mathcal{L}}{\partial\bm{h}}
  = \eta\!\left(\bm\psi_g - \sum_i p_i(t)\,\bm\psi_i\right),
\end{equation*}
which is Eq.~\eqref{eq:hdot}.
No external result is required. The equation follows from the chain
rule on the softmax and the definition of gradient flow.

\paragraph{Setup.}
Let $\lgap(t) = \langle\bm\psi_g,\bm{h}(t)\rangle - \langle\bm\psi_{c^*},\bm{h}(t)\rangle
= \langle\bm\psi_g - \bm\psi_{c^*},\,\bm{h}(t)\rangle$
denote the logit gap.
All embeddings are unit-normalised, with $\|\bm\psi_i\|=1$,
$G_{ij} = \bm\psi_i^\top\bm\psi_j$, $G_{gg}=G_{c^*c^*}=1$,
and $G_{gc^*} = \gmax$.

\paragraph{Exact time derivative of $\lgap$.}
Differentiating and substituting the gradient flow Eq.~\eqref{eq:hdot}
gives
\begin{equation}
  \dot\lgap(t)
  = \langle\bm\psi_g - \bm\psi_{c^*},\,\dot{\bm{h}}(t)\rangle
  = \eta\,\bigl\langle\bm\psi_g - \bm\psi_{c^*},\;
    \bm\psi_g - \textstyle\sum_i p_i(t)\,\bm\psi_i\bigr\rangle.
\end{equation}

Expanding the inner product gives
\begin{equation}
  \dot\lgap(t)
  = \eta\Bigl[\underbrace{\langle\bm\psi_g-\bm\psi_{c^*},\bm\psi_g\rangle}_{1-\gmax}
    \;-\;\sum_i p_i(t)\underbrace{\langle\bm\psi_g-\bm\psi_{c^*},\bm\psi_i\rangle}_{G_{gi}-G_{c^*i}}
    \Bigr].
  \label{eq:delta_dot_exact}
\end{equation}

\paragraph{Separating the $g$- and $c^*$-contributions.}
Splitting the sum into the $g$, $c^*$, and background terms gives
\begin{align}
  \sum_i p_i(G_{gi} - G_{c^*i})
  &= p_g(1 - \gmax) + p_{c^*}(\gmax - 1)
    + \sum_{i \neq g,c^*} p_i(G_{gi} - G_{c^*i}) \notag\\
  &= (p_g - p_{c^*})(1 - \gmax) + R_{\mathrm{bg}},
  \label{eq:sum_split}
\end{align}
where $R_{\mathrm{bg}} = \sum_{i \neq g,c^*} p_i(G_{gi}-G_{c^*i})$ is the
background drag term.

Substituting into Eq.~\eqref{eq:delta_dot_exact} gives
\begin{equation}
  \dot\lgap(t)
  = \eta(1-\gmax)\bigl[1 - (p_g - p_{c^*})\bigr]
    - \eta\,R_{\mathrm{bg}}.
  \label{eq:delta_dot_split}
\end{equation}

\begin{acknowledgments}
The authors acknowledge the use of Claude Sonnet 5 (Anthropic) for
assistance with numerical verification and manuscript consistency checking during
preparation of this manuscript. All AI-assisted content was critically
reviewed and verified by the authors. The authors thank Mahindra
University for supporting this research. The authors declare no
conflicts of interest.
\end{acknowledgments}

\section*{Data Availability Statement}
The code and data that support the findings of this study are
available from the corresponding author upon reasonable request.

\bibliographystyle{apsrev4-2}
\bibliography{refs}

@book{nielsen2000quantum,
  title     = {Quantum Computation and Quantum Information},
  author    = {Nielsen, Michael A and Chuang, Isaac L},
  year      = {2000},
  publisher = {Cambridge University Press}
}

@misc{sanh2019distilbert,
  title  = {DistilBERT, a distilled version of {BERT}: smaller, faster,
            cheaper and lighter},
  author = {Sanh, Victor and Debut, Lysandre and Chaumond, Julien and
            Wolf, Thomas},
  year   = {2019},
  eprint = {1910.01108},
  archivePrefix = {arXiv},
  primaryClass = {cs.CL}
}

@misc{black2021gptneo,
  title  = {{GPT-Neo}: Large Scale Autoregressive Language Modeling with
            Mesh-Tensorflow},
  author = {Black, Sid and Gao, Leo and Wang, Phil and Leahy, Connor and
            Biderman, Stella},
  year   = {2021},
  publisher = {Zenodo},
  doi    = {10.5281/zenodo.5297715}
}

@misc{allal2024SmolLM,
  title  = {SmolLM - blazingly fast and remarkably powerful},
  author = {Ben Allal, Loubna and Lozhkov, Anton and Bakouch, Elie and
            von Werra, Leandro and Wolf, Thomas},
  year   = {2024},
  howpublished = {\url{https://huggingface.co/blog/smollm}}
}

@inproceedings{yang2018breaking,
  title     = {Breaking the softmax bottleneck: A high-rank {RNN} language model},
  author    = {Yang, Zhilin and Dai, Zihang and Salakhutdinov, Ruslan and
               Cohen, William W},
  booktitle = {International Conference on Learning Representations},
  year      = {2018}
}

@inproceedings{press2017using,
  title     = {Using the output embedding to improve language models},
  author    = {Press, Ofir and Wolf, Lior},
  booktitle = {Proceedings of the 15th Conference of the European Chapter of the
               Association for Computational Linguistics},
  pages     = {157--163},
  year      = {2017}
}

@inproceedings{nanda2023progress,
  title     = {Progress measures for grokking via mechanistic interpretability},
  author    = {Nanda, Neel and Chan, Lawrence and Lieberum, Tom and
               Smith, Jess and Steinhardt, Jacob},
  booktitle = {International Conference on Learning Representations},
  year      = {2023}
}

@inproceedings{gao2019representation,
  title     = {Representation degeneration problem in training natural language
               generation models},
  author    = {Gao, Jun and He, Di and Tan, Xu and Qin, Tao and Wang, Liwei
               and Liu, Tie-Yan},
  booktitle = {International Conference on Learning Representations},
  year      = {2019}
}

@article{su2024roformer,
  title   = {{RoFormer}: Enhanced transformer with rotary position embedding},
  author  = {Su, Jianlin and Ahmed, Murtadha and Lu, Yu and Pan, Shengfeng and
             Bo, Wen and Liu, Yunfeng},
  journal = {Neurocomputing},
  volume  = {568},
  pages   = {127063},
  year    = {2024}
}

@inproceedings{li2021prefix,
  title     = {Prefix-tuning: Optimizing continuous prompts for generation},
  author    = {Li, Xiang Lisa and Liang, Percy},
  booktitle = {Proceedings of ACL-IJCNLP},
  pages     = {4582--4597},
  year      = {2021}
}

@article{brown2020language,
  title   = {Language models are few-shot learners},
  author  = {Brown, Tom and Mann, Benjamin and Ryder, Nick and Subbiah, Melanie and
             Kaplan, Jared D and Dhariwal, Prafulla and Neelakantan, Arvind and
             Shyam, Pranav and Sastry, Girish and Askell, Amanda and others},
  journal = {Advances in Neural Information Processing Systems},
  volume  = {33},
  pages   = {1877--1901},
  year    = {2020}
}

@inproceedings{devlin2019bert,
  title     = {{BERT}: Pre-training of deep bidirectional transformers for
               language understanding},
  author    = {Devlin, Jacob and Chang, Ming-Wei and Lee, Kenton and
               Toutanova, Kristina},
  booktitle = {Proceedings of NAACL-HLT},
  pages     = {4171--4186},
  year      = {2019}
}

@article{radford2019language,
  title   = {Language models are unsupervised multitask learners},
  author  = {Radford, Alec and Wu, Jeffrey and Child, Rewon and Luan, David and
             Amodei, Dario and Sutskever, Ilya},
  journal = {OpenAI Blog},
  volume  = {1},
  number  = {8},
  pages   = {9},
  year    = {2019}
}

@article{dubey2024llama3,
  title   = {The {Llama~3} Herd of Models},
  author  = {Dubey, Abhimanyu and Jauhri, Abhinav and Pandey, Abhinav and
             Kadian, Abhishek and Al-Dahle, Ahmad and Letman, Aiesha and others},
  journal = {arXiv preprint arXiv:2407.21783},
  year    = {2024}
}

@article{touvron2023llama,
  title   = {{LLaMA}: Open and efficient foundation language models},
  author  = {Touvron, Hugo and Lavril, Thibaut and Izacard, Gautier and
             Martinet, Xavier and Lachaux, Marie-Anne and Lacroix, Timoth{\'e}e and
             Rozi{\`e}re, Baptiste and Goyal, Naman and Hambro, Eric and
             Azhar, Faisal and others},
  journal = {arXiv preprint arXiv:2302.13971},
  year    = {2023}
}

@inproceedings{biderman2023pythia,
  title   = {Pythia: A suite for analyzing large language models across
             training and scaling},
  author  = {Biderman, Stella and Schoelkopf, Hailey and Anthony, Quentin and
             Bradley, Herbie and O'Brien, Kyle and Hallahan, Eric and Khan, Mohammad
             Aflah and Purohit, Shivanshu and Prashanth, USVSN Sai and Raff, Edward
             and others},
  booktitle = {International Conference on Machine Learning},
  pages     = {2397--2430},
  year      = {2023}
}

@article{ouyang2022training,
  title   = {Training language models to follow instructions with human feedback},
  author  = {Ouyang, Long and Wu, Jeffrey and Jiang, Xu and Almeida, Diogo and
             Wainwright, Carroll and Mishkin, Pamela and Zhang, Chong and
             Agarwal, Sandhini and Slama, Katarina and Ray, Alex and others},
  journal = {Advances in Neural Information Processing Systems},
  volume  = {35},
  pages   = {27730--27744},
  year    = {2022}
}

@inproceedings{hu2022lora,
  title     = {{LoRA}: Low-rank adaptation of large language models},
  author    = {Hu, Edward J and Shen, Yelong and Wallis, Phillip and
               Allen-Zhu, Zeyuan and Li, Yuanzhi and Wang, Shean and
               Wang, Lu and Chen, Weizhu},
  booktitle = {International Conference on Learning Representations},
  year      = {2022}
}

@inproceedings{houlsby2019parameter,
  title     = {Parameter-efficient transfer learning for {NLP}},
  author    = {Houlsby, Neil and Giurgiu, Andrei and Jastrzebski, Stanislaw and
               Morrone, Bruna and De Laroussilhe, Quentin and Gesmundo, Andrea and
               Attariyan, Mona and Gelly, Sylvain},
  booktitle = {International Conference on Machine Learning},
  pages     = {2790--2799},
  year      = {2019}
}

@article{dettmers2023qlora,
  title   = {{QLoRA}: Efficient finetuning of quantized {LLMs}},
  author  = {Dettmers, Tim and Pagnoni, Artidoro and Holtzman, Ari and
             Zettlemoyer, Luke},
  journal = {Advances in Neural Information Processing Systems},
  volume  = {36},
  year    = {2023}
}

@inproceedings{lester2021power,
  title     = {The power of scale for parameter-efficient prompt tuning},
  author    = {Lester, Brian and Al-Rfou, Rami and Constant, Noah},
  booktitle = {Proceedings of EMNLP},
  pages     = {3045--3059},
  year      = {2021}
}

@article{wei2022emergent,
  title   = {Emergent abilities of large language models},
  author  = {Wei, Jason and Tay, Yi and Bommasani, Rishi and Raffel, Colin and
             Zoph, Barret and Borgeaud, Sebastian and Yogatama, Dani and
             Bosma, Maarten and Zhou, Denny and Miculivicius, Donald and others},
  journal = {Transactions on Machine Learning Research},
  year    = {2022}
}

@article{power2022grokking,
  title   = {Grokking: Generalization beyond overfitting on small algorithmic datasets},
  author  = {Power, Alethea and Burda, Yuri and Edwards, Harri and Babuschkin,
             Igor and Misra, Vedant},
  journal = {arXiv preprint arXiv:2201.02177},
  year    = {2022}
}

@article{anderson1958absence,
  title   = {Absence of diffusion in certain random lattices},
  author  = {Anderson, Philip W},
  journal = {Physical Review},
  volume  = {109},
  number  = {5},
  pages   = {1492},
  year    = {1958}
}

@inproceedings{ethayarajh2019contextual,
  title   = {How contextual are contextualized word representations?
             Comparing the geometry of {BERT}, {ELMo}, and {GPT-2} embeddings},
  author  = {Ethayarajh, Kawin},
  booktitle = {Proceedings of EMNLP-IJCNLP},
  pages   = {55--65},
  year    = {2019}
}

@inproceedings{cai2021isotropy,
  title   = {Isotropy in the contextual embedding space: Clusters and functional
             roles},
  author  = {Cai, Xingyu and Dong, Jiaji and Rohatgi, Pratik and Church, Kenneth W},
  booktitle = {International Conference on Learning Representations},
  year    = {2021}
}

@article{coecke2020mathematical,
  title     = {Mathematical foundations for a compositional distributional
               model of meaning},
  author    = {Coecke, Bob and Sadrzadeh, Mehrnoosh and Clark, Stephen},
  journal   = {Linguistic Analysis},
  volume    = {36},
  pages     = {345--384},
  year      = {2010}
}

@inproceedings{zhang2018quantum,
  title   = {End-to-end quantum-like language models with application to
             question answering},
  author  = {Zhang, Peng and Niu, Jiabin and Su, Zhan and Wang, Benyou and
             Ma, Leyu and Song, Dawei},
  booktitle = {Proceedings of AAAI},
  volume  = {32},
  year    = {2018}
}

@inproceedings{vaswani2017attention,
  title     = {Attention is all you need},
  author    = {Vaswani, Ashish and Shazeer, Noam and Parmar, Niki and
               Uszkoreit, Jakob and Jones, Llion and Gomez, Aidan N and
               Kaiser, {\L}ukasz and Polosukhin, Illia},
  booktitle = {Advances in Neural Information Processing Systems},
  volume    = {30},
  year      = {2017}
}

@inproceedings{maynez2020faithfulness,
  title     = {On faithfulness and factuality in abstractive summarization},
  author    = {Maynez, Joshua and Narayan, Shashi and Bohnet, Bernd and
               McDonald, Ryan},
  booktitle = {Proceedings of ACL},
  pages     = {1906--1919},
  year      = {2020}
}

@article{ji2023survey,
  title   = {Survey of hallucination in natural language generation},
  author  = {Ji, Ziwei and Lee, Nayeon and Frieske, Rita and Yu, Tiezheng and
             Su, Dan and Xu, Yan and Ishii, Etsuko and Bang, Ye Jin and
             Madotto, Andrea and Fung, Pascale},
  journal = {ACM Computing Surveys},
  volume  = {55},
  number  = {12},
  pages   = {1--38},
  year    = {2023}
}

@inproceedings{intrinsic2020,
  title   = {Intrinsic dimensionality explains the effectiveness of language
             model fine-tuning},
  author  = {Aghajanyan, Armen and Zettlemoyer, Luke and Gupta, Sonal},
  booktitle = {Proceedings of ACL},
  year    = {2021}
}

@inproceedings{mosbach2021stability,
  title   = {On the stability of fine-tuning {BERT}: Misconceptions, explanations,
             and strong baselines},
  author  = {Mosbach, Marius and Andriushchenko, Maksym and Klakow, Dietrich},
  booktitle = {International Conference on Learning Representations},
  year    = {2021}
}

@article{carson2025statistical,
  title   = {A Statistical Physics of Language Model Reasoning},
  author  = {Carson, Jack David and Reisizadeh, Amir},
  journal = {arXiv preprint arXiv:2506.04374},
  year    = {2025}
}

@article{dodge2020finetuning,
  title   = {Fine-tuning pretrained language models: Weight initializations,
             data orders, and early stopping},
  author  = {Dodge, Jesse and Ilharco, Gabriel and Schwartz, Roy and Farhadi,
             Ali and Hajishirzi, Hannaneh and Smith, Noah},
  journal = {arXiv preprint arXiv:2002.06305},
  year    = {2020}
}

@inproceedings{kumar2022fine,
  title   = {Fine-tuning can distort pretrained features and underperform
             out-of-distribution},
  author  = {Kumar, Ananya and Raghunathan, Aditi and Jones, Robbie and Ma,
             Tengyu and Liang, Percy},
  booktitle = {International Conference on Learning Representations},
  year    = {2022}
}

@article{lecun2022path,
  title   = {A Path Towards Autonomous Machine Intelligence},
  author  = {LeCun, Yann},
  journal = {OpenReview},
  year    = {2022},
  note    = {Version 0.9.2}
}

\end{document}